\pdfoutput=1

\documentclass{bmcart}

\usepackage{hyperref}
\usepackage{algorithmicx}
\usepackage[noend]{algpseudocode}
\usepackage{algorithm}
\usepackage{graphicx}
\usepackage{amsmath}

\usepackage{multirow}
\usepackage{acro}

\acsetup{first-style=short}

\DeclareAcronym{wss}{
  short = WSS,
  long  = working set selection,
  class = abbrev
}

\DeclareAcronym{smo}{
  short = SMO,
  long  = sequential minimal optimization,
  class = abbrev
}

\DeclareAcronym{vc_dim}{
  short = VC dimension,
  long  = Vapnik Chervonenkis dimension,
  class = abbrev
}

\DeclareAcronym{flann}{
  short = FLANN,
  long  = fast library for approximate nearest neighbors,
  class = abbrev
}

\DeclareAcronym{ann}{
  short = ANN,
  long  = approximate nearest neighbors,
  class = abbrev
}

\DeclareAcronym{sd}{
  short = SD,
  long  = standard deviation,
  class = abbrev
}

\DeclareAcronym{knng}{
  short = K-NNG,
  long  = K-nearest neighbors graph,
  class = abbrev
}

\DeclareAcronym{sns}{
  short = SNS ,
  long  = superficial neighbor search,
  class = abbrev
}

\DeclareAcronym{ens}{
  short = ENS ,
  long  = exclusive neighbor search,
  class = abbrev
}

\DeclareAcronym{gsh}{
  short = GSH ,
  long  = graph shedding heuristic,
  class = abbrev
}

\DeclareAcronym{gch}{
  short = GCH ,
  long  = graph coarsening heuristic,
  class = abbrev
}

\DeclareAcronym{nhs}{
  short = NHS ,
  long  = nearest hypothesis search,
  class = abbrev
}

\DeclareAcronym{numOfPoints}{
  short = $n$ ,
  long  = the number of data points,
  sort  = n ,
  class = nomencl
}

\DeclareAcronym{dimensions}{
  short = $d$ ,
  long  = the number of features,
  sort  = d ,
  class = nomencl
}

\DeclareAcronym{numOfClusters}{
  short = $n_{c}$ ,
  long  = the number of initial clusters (input parameter),
  sort  = nc ,
  class = nomencl
}

\DeclareAcronym{wij}{
  short = $w_{i,j}$ ,
  long  = weight of \ensuremath{(i, j)} edge,
  sort  = wij ,
  class = nomencl
}

\DeclareAcronym{tci}{
  short = $\text{tc}(i)$ ,
  long  = target class of the \ensuremath{i^{th}} node,
  sort  = tci ,
  class = nomencl
}

\DeclareAcronym{ci}{
  short = $C_{I}$ ,
  long  = constant for internal classification pattern,
  sort  = ci ,
  class = nomencl
}

\DeclareAcronym{ce}{
  short = $C_{E}$ ,
  long  = constant for external classification pattern,
  sort  = ce ,
  class = nomencl
}

\DeclareAcronym{r}{
  short = $R$ ,
  long  = reach constant (input parameter),
  sort  = r ,
  class = nomencl
}

\DeclareAcronym{ri}{
  short = $r_{i}$ ,
  long  = reach of the \ensuremath{i^{th}} node,
  sort  = ri ,
  class = nomencl
}

\DeclareAcronym{xi}{
  short = $x_{i}$ ,
  long  = position of the \ensuremath{i^{th}} node,
  sort  = xi ,
  class = nomencl
}

\DeclareAcronym{xj}{
  short = $x_{j}$ ,
  long  = position of the \ensuremath{j^{th}} node,
  sort  = xj ,
  class = nomencl
}

\DeclareAcronym{ni}{
  short = $N_{i}$ ,
  long  = set of neighbors of \ensuremath{i^{th}} node,
  sort  = ni ,
  class = nomencl
}

\DeclareAcronym{cg}{
  short = \ensuremath{\text{cost}(G)} ,
  long  = cost of graph \ensuremath{G},
  sort  = cg ,
  class = nomencl
}

\DeclareAcronym{e}{
  short = $e$ ,
  long  = an edge,
  sort  = e ,
  class = nomencl
}

\DeclareAcronym{Ep}{
  short = $E^{'}$ ,
  long = subset of $E$ (set of all edges) such that \ensuremath{ \{e \mid e \epsilon E \mbox{, and $>$ GC edge cut}\} },
  sort  = ep ,
  class = nomencl
}

\DeclareAcronym{we}{
  short = $w_{e}$ ,
  long  = weight of an edge $e$,
  sort  = we ,
  class = nomencl
}

\newcommand\AlgphaseNotext[1]{
\vspace*{-.7\baselineskip}\Statex\hspace*{\dimexpr-\algorithmicindent-2pt\relax}\rule{\textwidth}{0.4pt}
}

\algdef{SE}[EVENT]{Event}{EndEvent}[1]{\textbf{upon event}\ #1\ \algorithmicdo}{\algorithmicend\ \textbf{event}}%
\algtext*{EndEvent}

\algdef{SE}[GEN_EVENT]{GenEvent}{EndGenEvent}[1]{\textbf{trigger event}\ #1\ }{\algorithmicend\ \textbf{gen event}}%
\algtext*{EndGenEvent}

\begin{document}

\begin{frontmatter}

\begin{fmbox}
\dochead{Methodology}

\title{A graphical heuristic for reduction and partitioning of large datasets for scalable supervised training}

\author[
   addressref={aff1},                   % id's of addresses, e.g. {aff1,aff2}
   corref={aff1},                       % id of corresponding address, if any
   email={sumedhyadav.iitkgp@gmail.com}   % email address
]{\inits{SY}\fnm{Sumedh} \snm{Yadav}}

\address[id=aff1]{%                           % unique id
  \orgname{Gstech Technology Pvt. Ltd.}, % university, etc
  \street{415, 2nd Floor, 16th Cross Road, 17th Main Road, HSR Layout Sector 4}, %
  \postcode{560102},                         % post or zip code
  \city{Bengaluru},                              % city
  \cny{India}                                    % country
}

\author[
   addressref={aff2},                   % id's of addresses, e.g. {aff1,aff2}
   email={m.bode@itv.rwth-aachen.de}   % email address
]{\inits{MB}\fnm{Mathis} \snm{Bode}}

\address[id=aff2]{%                           % unique id
  \orgname{Institute for Combustion Technology, RWTH Aachen University}, % university, etc
  \street{Templergraben 64}, %
  \postcode{52056},                         % post or zip code
  \city{Aachen},                              % city
  \cny{Germany}                                    % country
}

\end{fmbox}% comment this for two column layout

\begin{abstractbox}

\begin{abstract} % abstract

A scalable graphical method is presented for selecting, and partitioning datasets for the training phase of a classification task. For the heuristic, a clustering algorithm is required to get its computation cost in a reasonable proportion to the task itself. This step is proceeded by construction of an information graph of the underlying classification patterns using approximate nearest neighbor methods. The presented method constitutes of two approaches, one for reducing a given training set, and another for partitioning the selected/reduced set. The heuristic targets large datasets, since the primary goal is significant reduction in training computation run-time without compromising prediction accuracy. Test results show that both approaches significantly speed-up the training task when compared against that of state-of-the-art shrinking heuristic available in LIBSVM. Furthermore, the approaches closely follow or even outperform in prediction accuracy. A network design is also presented for the partitioning based distributed training formulation. Added speed-up in training run-time is observed when compared to that of serial implementation of the approaches.

\end{abstract}

\end{abstractbox}
\end{frontmatter}

\section*{Introduction}
Two decades earlier, some of the most seminal works in machine learning were done on training set selection \cite{AlonYLevy, AvrimLBlum} under the banner of relevance reasoning. However, the better part of recent works have been exclusively towards feature selection \cite{Weng2017, guyon2008feature}. With increased processing power, run time of training is feasible even for datasets erstwhile considered large. Additionally, dimensionality (\ac{dimensions}) dominates dataset size (\ac{numOfPoints}) in the algorithmic complexities of learning algorithms. In the training phase, less data points mean fewer generalization guarantees, however, as we are moving in the era of big data, even the fastest classification algorithms are taking un-feasible time to train models. When data sources are abundant, it is befitting to separate data based on relevance to the learning task. This has led to a renewed interest in the once famous problem statement of relevance reasoning \cite{HatemAFayed,CodyColeman}. Reasoning on relevance to get improved scalability of classification algorithms is currently explored on graphical/network data \cite{AndreasLoukas}, and learned models \cite{Weinberg2019}.

One research area where training set selection has been given attention to is support vector machines (SVM). Generally, these selection methods can be divided into two types. The first type of methods aims to modify the SVM formulation so that it can be applied to large datasets. Many approaches have worked successfully in the past, including sequential minimal optimization (\ac{smo}) \cite{PaiHsuenChen, RongEnFan}, and genetic programming \cite{JakubNalepa}. The first type of methods, however, do not benefit from reducing the size of training data because they only deal with data handling. Reduction of data size is the quintessential advantage of the second kind of methods \cite{StanSalvador,MamounAwad,JairCervantesa,XiaoouLi,JigangWang,MichaelEMavroforakis,GlennFung,PredragNeskovic,HwanjoYu}. These methods focus on segregating the data points on relevance to the classification task. Computation time is reduced by actively reducing \ac{numOfPoints}. Handling data was the central theme of research during the mid-2000s instead of reducing it. Despite the apparent advantage of reduction of data, researchers made sincere efforts for methods of the first type \cite{PaiHsuenChen,RongEnFan,JakubNalepa,LIBSVM} in comparison to the latter type. For instance, formulations implemented in LIBSVM \cite{LIBSVM}, which is a widely used benchmark library for SVM methods, are of the first type of methods.

Few existing works of the second type are limited in one respect or other. For instance, clustering based SVM (CB-SVM) work by Yang et al. \cite{HwanjoYu}, and others \cite{MamounAwad, StanSalvador} which resulted in huge speed-ups are limited to linear kernels only. A geometric approach of minimum ball enclosing by Cervantes et al. \cite{JairCervantesa} requires two stages of SVM training. A similar method by Li et al. \cite{XiaoouLi} suffers from a random selection of data. They have reported 92.0\% prediction accuracy for a separable dataset for which LIBSVM reproduces 99.9\%. 

The presented approach also falls under the second type. As will be shown, the presented selection scheme is very deterministic in prediction accuracy. A model trained with the presented heuristic results in close or better prediction accuracy than that of full data training. One recurring problem with the second type of methods is inefficient space searching scalability \cite{XiaoouLi, JigangWang, MichaelEMavroforakis, GlennFung, PredragNeskovic}. It becomes worse for high dimensional data, where the heuristic takes more time than training itself \cite{AsdrubalLopezChau}. This issue is addressed by the use of state-of-the-art approximate nearest neighbor (\ac{ann}) methods \cite{MariusMuja} which are highly scalable.

In LIBSVM, SMO decomposition method of Fan et al. \cite{RongEnFan} is available for the classification task. Training set selection compares in principle to working set selection (\ac{wss}) in the context of SMO or similar decomposition methods. Furthermore, a shrinking technique is available for these formulations to remove bounded components during iterations, effectively reducing the optimization problem \cite{RongEnFan}. As will be shown, data selection in the presented heuristic only depends on the underlying classification patterns, giving it an essential advantage of generic applicability to the majority of classification algorithms, including the SMO formulations of LIBSVM. For these reasons, the state-of-the-art shrinking heuristic of LIBSVM is compared to the presented heuristic.

The presented heuristic augments clustering based approaches \cite{HwanjoYu, MamounAwad, StanSalvador} by constructing an (approximate) information graph out of the clustered data. This graph acts as a proxy for reducing the training set. A novel edge weight scheme captures the underlying classification patterns in the graph. The graph is then pruned via filtering on the edge weights to select a relevant dataset that can be used for the training task. Furthermore, a graph coarsening approach is presented to break the selected/reduced set into further partitions that are independently available for training, leading to an approximate learning scheme. Both the methods lead to reduction in the number of training data points, which reduces complexity of the training algorithm, giving performance advantages.

Most of the existing methods of both types are limited to the SVM class of algorithms \cite{StanSalvador,MamounAwad,JairCervantesa,XiaoouLi,JigangWang,MichaelEMavroforakis,GlennFung,PredragNeskovic,HwanjoYu,PaiHsuenChen,RongEnFan,JakubNalepa,LIBSVM}. Generic applicability on a majority of classification algorithms is another advantage of the presented heuristic. It gives an opportunity to use the heuristic as a pre-processing tool, separate from the classification algorithm. Since the data points are selected based on their relevance to the classification task, the resultant reduced training set is much more balanced in size across the target classes. In other words, the formulation addresses the problem statement of class imbalance, which is a topic of current research in big data \cite{Leevy2018}.

The remainder of this paper is organized as follows, the `Methodology' section describes formulations of the proposed heuristic in detail. The heuristic is evaluated with a number of tests, and datasets in the `Results' section. Finally, the `Conclusions' section summarizes concluding remarks, and ideas for future work in the heuristic formulation.

\section*{Methodology}
The heuristic procedure organically divides into the following steps,
\begin{itemize}
\item Clustering step
\item Graph knitting scheme
\item Graph shedding scheme
\item Graph clubbing scheme
\item Pre-processor for the testing phase
\end{itemize}

The training phase proceeds after the first four steps, whereas the testing phase follows the last step of the heuristic formulation. The clustering step is used to get a computationally feasible resolution of the underlying data. A weighted graph is constructed next in the graph knitting scheme, using a three parts algorithm, and an edge weight scheme, which captures the classification patterns completely. Significant nodes of this graph with respect to the classification task are determined in the graph shedding step. Finally, the graph clubbing step divides these nodes into partitions that can be trained independently using a directional aspect of the graph coarsening objective achieved via another three parts algorithm. Because of the multiple data partitions which translate into as many classifiers, there is a need to determine which classifier to choose for testing a data point. This is achieved by the pre-processor for the testing phase. Lastly, a network application is designed which distributes the obtained partitions in a load-balancing, and a communication-free manner.

From a computational point of view, run-time profile of the first four steps for a typical run case ($n$ $\approx$100k, $d$ = 2, $n_c$ $\approx$350 clusters) is,

\begin{itemize}
\item Clustering step - $\approx$750 ms or $>$98\%
\item Graph knitting scheme - $\approx$10 ms or 1\%
\item Graph shedding, and graph clubbing schemes - $\approx$2 ms ($< 1\%$)
\end{itemize}

On the other hand, the pre-processor for the testing phase takes $\approx$5\% of the testing phase time. The clustering step is predominant over subsequent steps with a run-time complexity in $n$, whereas that of all the other steps is in the order of number of clusters (\ac{numOfClusters}). The input parameter to the clustering step, $n_{c}$, controls the granularity of data representation. Typically, the ratio of $n$ to that of $n_{c}$, which is also known as nominal \ac{vc_dim}, is 10 to 300, explaining the run-time dominance of the clustering step. The graph knitting scheme becomes the most computation-intensive upon considering the subsequent steps, because it involves heavy space searching. It is to be noted that the run-time percentage profile can vary a lot depending on the nominal VC dimension or the input parameter $n_{c}$.

\subsection*{Clustering step}
The clustering step is used to lower the resolution of the underlying data. In principle, the presented heuristic does not require this step. However, it is not computationally feasible to execute the subsequent steps with a run-time complexity in $n$ instead of $n_{c}$. Additionally, it will also affect the generic applicability of the heuristic, which is discussed later in the `Graph shedding scheme' step.

The step consists of a standard K-means++ \cite{KMeans++} clustering algorithm, and a metric to store classification patterns of the original data. K-means++ provides improved initial seeding of clusters over the traditional K-means method. This choice leads to running the clustering algorithm for a nominal number of iterations, typically 5. In the current implementation, every cluster center maintaines the target class through the weighted average calculation over all its data points. However, advanced metrics can be constructed to unearth more characteristics of patterns from the clustered data.

Although this step serves only for coarsening the data representation, it dominates the computation cost of the first four steps. Two state-of-the-art K-means++ implementations were tested, K-MeansRex \cite{GITHUBKmeansRex}, and scalable mlpack package \cite{MLPACK}. For a test run with the data points in the range of 1 to 100K ($d$ = 2, $n_c$ = 100), K-means++ from mlpack was $1.82$ times faster on average in execution than K-MeansRex's implementation. Therefore, the K-means++ from mlpack is chosen as the standard clustering algorithm in this work.

Given the vast research literature available for clustering methods, there are other implementations available. One of such improvements would be the scalable K-means++ by Bahmani et al. \cite{bahmani}, which is shown to be considerably faster than native K-means++. Another practical option is to exploit K-means implementations from a proven distributed computing platform \cite{Spark}.

\subsection*{Graph knitting scheme}
From this step onwards, the presented heuristic digresses from most of the existing geometric approaches of the second type, primarily because of the choice of graph to represent the classification dataset, and the use of seminal works in neighbor searching methods. First, the choice of a weighted graph opened the possibility of using well researched work on graph coarsening, which is the foundation of the graph clubbing step. Second, most of the existing approaches could not benefit from seminal works in the neighbor searching methods, which have contributed profoundly to the success of computer vision. The fast library for approximate nearest neighbors (FLANN) search engine \cite{MariusMuja} is used in this work.

An information graph is constructed once a reasonable representation of the underlying data is obtained. Two major challenges include determination of neighbors, and capturing the classification patterns in edge weights. First, neighbors are determined such that the whole hypothesis is covered while passively enforcing regularity and planarity in the graph. The neighbors are determined in two steps, superficial search, and exclusive search, presented in Part \hyperref[algo:Algorithm1PartI]{I}, and \hyperref[algo:Algorithm1PartI]{II} of Algorithm I, respectively. Part \hyperref[algo:Algorithm1PartI]{III} of the algorithm controls skewness of the graph. Second, a two-fold pattern capturing edge weight scheme is presented in Eq. \hyperref[eqn:PowerScheme]{2}.

The first challenge of neighbor determination is addressed in two stages, superficial neighbor search (\ac{sns}), and exclusive neighbor search (\ac{ens}), presented in Part \hyperref[algo:Algorithm1PartI]{I}, and \hyperref[algo:ens]{II}, respectively. In the algorithm, the number of neighbors is nominally controlled by an input parameter $nn$, the number of desired neighbors. For every $i^{\text{th}}$ node, variable $no\_tot\_neigh[i]$ in Line 1 of Part I is used to track the size of the neighbor list.

However, neighbor list for the \ensuremath{i^{\text{th}}} node can be terminated before adding $nn$ neighbors by updating the variable $neigh\_list\_finish[i]$ to {\scriptsize TRUE} in Line 4 of Part I. By limiting the number of neighbors in this way, construction of the graph can be controlled.

Part \hyperref[algo:Algorithm1PartI]{I} is used to look for $nn$ nearest neighbors, regardless of the target class. The algorithm takes input an empty graph, $G(V, \phi)$, which is formed with the set of the cluster centers, $V$, after the clustering step. This set is the search space passed to FLANN space indexing utility in Line 8, and 9. The input constant $R$ is explained later in Eq. \hyperref[eqn:reach]{1}. Objective of the part is to fill the graph, $G$, with $E_I$ set of edges. This is similar to the construction of a K-nearest neighbor graph (\ac{knng}). However, an input parameter, {\scriptsize MAX\_SAME\_CLASS\_NEIGH}, is used to limit the number of same class neighbors in Line 13, so that the remainder of edges for the $i^{\text{th}}$ node, $nn - no\_tot\_neigh[i]$, are constructed for nodes with opposite target class. Nodes for which all $nn$ neighbors are found will vary depending on characteristics of the data. They will be excluded from the next parts. An additional computation of reach, in Line 15, is maintained for every node. The metric presented in Eq. \hyperref[eqn:reach]{1} is similar to the Hausdorff distance \cite{PredragNeskovic}. The distance utility of FLANN, in Line 10, is used to compute the summation in Eq. \hyperref[eqn:reach]{1}, whereas a scaling constant ($R$) controls the reach according to
\begin{eqnarray}
r_i = R \times \sum_{j\epsilon N_i} |x_i - x_j|,
\label{eqn:reach}
\end{eqnarray}

where \ac{r} is the scaling constant, \ac{ri} is reach of the $i^{\text{th}}$ node, \ac{xi} is position of the $i^{\text{th}}$ node, \ac{xj} is position of the $j^{\text{th}}$ node, and \ac{ni} denotes set of same class neighbors for the $i^{\text{th}}$ node.

% =================New=================
\begin{algorithm}
\label{algo:Algorithm1PartI}
\caption{\textbf{1, Part I :} Superficial Neighbor Search}
\begin{flushleft}
Input : $G(V,\phi)$, $nn$, $R$, {\scriptsize{MAX\_SAME\_CLASS\_NEIGH}}\\
Output : $G(V,E_I)$ - $E_I$ is the set of edges, stored in $neigh\_list[]$
\end{flushleft}
\begin{algorithmic}[1]
\AlgphaseNotext{}
\State $no\_tot\_neigh[ ] \gets \phi$
\State $no\_same\_class\_neigh[ ] \gets \phi$
\State $neigh\_list[] \gets \phi$ \algorithmiccomment{neighbor list is NULL for each $i~\epsilon~V$}
\State $neigh\_list\_finish[] \gets \text{\scriptsize{FALSE}}$
\State $reach[ ]\gets \phi$
\AlgphaseNotext{}
\State $indices[ ][ ]\gets \phi$
\State $dists[ ][ ]\gets \phi$
\State $flann()\gets V$ \algorithmiccomment {FLANN instance with all nodes}
\State $flann.index()$
\State $indices, dists \gets flann.ann\_search(nn) $ \algorithmiccomment {ANN search}
\AlgphaseNotext{}
\ForAll {$i~\epsilon~V$} 
  \ForAll {$j~\epsilon~indices[i]$}
    \If {$same\_tc(i,j)$ and \text{$no\_same\_class\_neigh[i]< \text{\scriptsize{MAX\_SAME\_CLASS\_NEIGH}}$}}
      \State $neigh\_list[i]\,{+}{=}\,j$\algorithmiccomment {add $j$ to neighbor list of $i$}
      \State $reach[i]\,{+}{=}\,dists[i][j]$\algorithmiccomment {update reach of $i$}
      \State $no\_tot\_neigh[i]\,{+}{+}$
      \State $no\_same\_class\_neigh[i]{+}{+}$
    \EndIf
    \If {not $same\_tc(i,j)$}
      \State $neigh\_list [i]\,{+}{=}\,j$
      \State $no\_tot\_neigh[i]\,{+}{+}$
    \EndIf
  \EndFor
  \State $reach [i] \, {\times}{=} \, R$
  \If {$no\_tot\_neigh[i]\,{=}\,nn$}
    \State $neigh\_list\_finish[i] \gets \text{\scriptsize{TRUE}}$ \algorithmiccomment {neighbor searching finished for $i$}
  \EndIf
\EndFor
\end{algorithmic}
\end{algorithm}

In order to capture the classification patterns completely, it is necessary to make edges along the hypothesis of the classification data. So, Part \hyperref[algo:ens]{II} extends neighbor searching exclusively for nodes of the opposite target class. The $i^{\text{th}}$ node is considered only if there is a remainder requirement of neighbors, \ensuremath{nn_{r}[i]}, where \ensuremath{nn_{r}[i] = nn - no\_tot\_neigh[i]}. In the part, nodes of class 2 forms the search space in Line 3, in which neighbors are searched for class 1 nodes. This step along with the vice-versa case forms one iteration of the ENS procedure. For each iteration, a search space of the opposite target class is constructed in Line 3.

\begin{algorithm}
\label{algo:ens}
\caption{\textbf{1, Part II :} Exclusive (other class) Neighbor Search}
\begin{flushleft}
Input : $G(V,E_I$), class\_1\_nodes, class\_2\_nodes, $nn_{r}$[], reach[], {\scriptsize{NEIGH\_LIMIT}}\\
Output : $G(V,E_{II})$ - updated graph with $E_{II} - E_{I}$ set of new edges, updated in $neigh\_list[]$
\end{flushleft}
\begin{algorithmic}[1]
\AlgphaseNotext{}
\State $class\_1\_indices[][]\gets \phi$
\State $class\_1\_dists[][]\gets \phi$
\State $flann() \gets class\_2\_nodes$ \algorithmiccomment {FLANN instance with only class 2 nodes}
\State $flann.index()$
\State $class\_1\_indices, class\_1\_dists \gets flann.ann\_search(nn_{r}, class\_1\_nodes)$ \algorithmiccomment {ANN search}
\AlgphaseNotext{}
\State $node\_neigh[ ]\gets \phi$
\ForAll {$ i\,\epsilon \,class\_1\_nodes\,\text{and} $ not $ neigh\_list\_finish[i]$}
  \ForAll {$ j\text{ }\epsilon\,class\_1\_indices[i] $}
    \If {$node\_neigh[j] < \text{\scriptsize{NEIGH\_LIMIT}}$ and $reach[i] > class\_1\_dists[i][j]$}
      \State $neigh\_list(i)\,{+}{=} \,j$ \algorithmiccomment {add $j$ to neighbor list of $i$}
      \State $no\_tot\_neigh[i]\,{+}{+}$
      \State $node\_neigh[j]\,{+}{+}$
    \EndIf
    \If {$class\_1\_dists[i][j] > reach[i]$}
      \State $neigh\_list\_finish[i] \gets \text{\scriptsize{TRUE}}$ \algorithmiccomment {$i$ doesn't belong to convex hull}
    \EndIf
  \EndFor
\EndFor
\end{algorithmic}
\end{algorithm}

Inclusion of a neighbor after space searching in Line 5 is stringent compared to SNS. An input parameter, {\scriptsize NEIGH\_LIMIT}, and the computed reach are used to limit the availability of node $j$ as prospect neighbor in Line 9. For every $j^{\text{th}}$ node added as a neighbor, the variable $node\_neigh[j]$, initiated in Line 6, is incremented. Reach is used to further update the boolean $neigh\_list\_finish[i]$ for node $i$, even if \ensuremath{nn - no\_tot\_neigh[i] > 0} in Line 13. Such an \ensuremath{i^{\text{th}}} node tends to be an internal node of a target class. In other words, reach only encourages the convex hull nodes of one class to choose neighbors with nodes of the opposite class, aiding in planar construction of the graph.

$node\_neigh$ counter is required in Part II to avoid a node that might habitually come up as a prospect neighbor despite not being very representative of a target class. It otherwise leads to a skewed graph concerning node degree. $node\_neigh$ in Line 3 of Part \hyperref[algo:Algorithm1PartIII]{III} is used to reduce the search space of every target class, controlling skewness of the constructed graph.

\begin{algorithm}
\label{algo:Algorithm1PartIII}
\caption{\textbf{1, Part III :} Search/Indexing Space Reduction}
\begin{flushleft}
Input : $class\_1\_nodes$, $node\_neigh[]$, {\scriptsize{NEIGH\_LIMIT}}\\
Output : $class\_1\_nodes$
\end{flushleft}
\begin{algorithmic}[1]
\AlgphaseNotext{}
\State $new\_class\_1\_nodes \gets \phi $
\ForAll {$ i\,\epsilon \,class\_1\_nodes$}
  \If {$node\_neigh[i] < \text{\scriptsize{NEIGH\_LIMIT}}$}
    \State $new\_class\_1\_nodes\, {+}{=}\,i$ \algorithmiccomment {$i$ added in new nodes list}
  \EndIf
\EndFor
\State $class\_1\_nodes \gets new\_class\_1\_nodes$
\end{algorithmic}
\end{algorithm}

The second challenge is to capture the classification patterns in edge weights, for which a two-fold edge weight scheme is designed. First, each node measures its internal pattern as the absolute difference from one of the two target classes. Second, every pair of nodes in an edge measures external pattern by the relative difference of their target class. Individual contributions are added via the power scheme in Eq. \hyperref[eqn:PowerScheme]{2} to weigh the edge according to
\begin{eqnarray}
w_{i,j} = C_I^{1-|\text{tc}(i)|} + C_I^{1-|\text{tc(j)}|} + C_E^{|\text{tc}(i)-\text{tc}(j)|},
\label{eqn:PowerScheme}
\end{eqnarray}

where \ac{wij} is weight of the \ensuremath{(i, j)^{\text{th}}} edge, \ac{tci} denotes target class of the $i^{\text{th}}$ node, \ac{ci}, and \ac{ce} are constants for internal, and external classification patterns, respectively, and quantities in $||$ are absolute values.

The use of state-of-the-art implementation like FLANN, for neighbor searching cannot be overemphasized. For instance, in a typical run case ($n_{c} > 300$ and $nn > 3$), approximate neighbor searching method of FLANN was $\approx$1000 times faster when compared to the exact algorithm for nearest neighbor searching, which involves for every node, computing distances with all the remaining nodes, and then sorting them to determine the nearest neighbors. The run-time advantage is clear when comparing the complexity of exact graph construction, $O(n^3\text{log}(n))$, to that of approximate methods offered by FLANN \cite{MariusMuja}.

\subsection*{Graph shedding scheme}
Once a weighted graph is obtained, an edge cut based filtering presented in Algorithm~\hyperref[algo:Algorithm2]{2} separates the training dataset into relevant, and non-relevant. For every node, the neighbor list is iterated to check for a significant edge in Line 5, and when found, that node is added to $relevant\_node\_list$ in Line 11. This leads to a training set selection that the second type of approach aims to achieve. Note that the result of this step depends on the characteristics of the graph, such as how well connected the graph is, and how well the underlying classification patterns have been captured. Algorithm 1 with the edge weight scheme address these issues.

\begin{algorithm}
\label{algo:Algorithm2}
\caption{\textbf{2 :} Graph Shedding}
\begin{flushleft}
Input :  $G(V,E)$, \text{\scriptsize{EDGE\_CUT}}\\
Output : $relevant\_node\_list[]$
\end{flushleft}
\begin{algorithmic}[1]
\AlgphaseNotext{}
\State $relevant\_node\_list[] \gets \phi $
\ForAll {$i \,\epsilon \, V$}
  \State $critical \gets \text{\scriptsize{FALSE}}$
  \ForAll {$j\,\epsilon \, neigh\_list[i]$}
    \If {$ w_{i,j}\,>=\,\text{\scriptsize{EDGE\_CUT}}$}
      \State $critical \gets \text{\scriptsize{TRUE}}$
      \State $break$
    \EndIf
  \EndFor
\If {not $critical$}
  \State $G\,{-}{=}\,i$\algorithmiccomment {remove the $i^{th}$ node from $G$}
\Else 
\State {$relevant\_node\_list\,{+}{=}\,i$}\algorithmiccomment {add $i$ to potential critical list}
\EndIf
\EndFor
\end{algorithmic}
\end{algorithm}

The role of nominal VC dimension drips down in the pruned graph as well. Since it controls the granularity of the underlying data, it also controls the granularity of data selection. That gives the heuristic an essential advantage in terms of limiting data shrinkage while selecting the relevant data points. Because the selection is done via clusters, and the ratio of $n$ to $n_{c}$ is typically $> 10$, meaning that many data points that are not very close to the hypothesis boundary of the classification patterns are also selected. That gives an extra buffer of data points upon which another selection method of both types applies. The majority of classification algorithms, for example neural methods, gaussian processes et cetera, can use the heuristic. Until this step, the presented heuristic's aim matches with the existing approaches of the first, and second type. For comparison purposes, the heuristic until this step (including) is referred to as \ac{gsh} for the remainder of this work. Edge cut for GSH is referred to as GS edge cut.

\subsection*{Graph clubbing scheme}
\subsubsection*{Formulation}
This step extends the problem statement of training set selection to further breaking the reduced training set into few partitions or critical chunks, each of which can be trained independently by virtue of Part~\hyperref[algo:Algorithm3PartI]{I} to~\hyperref[algo:Algorithm3PartIII]{III} of Algorithm~\hyperref[algo:Algorithm3PartI]{3}. The main aim is to design an approximate formulation that is theoretically faster, even for serial execution. The algorithm divides the training set into few partitions such that the number of computations are reduced significantly in the training phase. Consider that the order of complexity of most of the classification algorithms is higher than linear, that is $a > 1$ if $O(n^{a})$ is the complexity. Now, the graph clubbing scheme doesn't change the order, however it results in significant reduction in total computations. For example, for a classification algorithm with $O(n^{2})$ complexity, if $C \times n^{2}$, where $C$ is a constant, is the original number of computations; then after data reduction by the graph clubbing scheme, there are four equal sized partitions/critical chunks. Now the number of computations is $C \times (4 \times (\frac {n}{4})^2 ) = C \times \frac{n^2}{4}$ or a quarter of the original number.

Independence during the training phase of each obtained partition is mainly because of the directional aspect of the algorithm, which is achieved via the edge weight scheme. The directional aspect is responsible for two objectives, namely obtaining equally-sized partitions, and ensuring orthogonality of the hypothesis boundary with neighboring partitions' boundary. Two ways in which the edge weight scheme is leveraged for the directional aspect is in the priority aspect of the partial weighted matching (PWM) algorithm, Part~\hyperref[algo:Algorithm3PartI]{I}, and the re-assessment aspect of the coarsening formulation, Part~\hyperref[algo:Algorithm3PartII]{II}. The graph clubbing algorithm, Part~\hyperref[algo:Algorithm3PartIII]{III} ties Part I, and II in an iterative scheme. Each of such obtained partitions can now be trained independently, giving further leverage for a nominal number of worker processes.

The partial weighted matching (PWM), Part~\hyperref[algo:Algorithm3PartI]{I}, is designed for the weighted graph obtained from the graph knitting step. The obtained matching is partial because edges weighing less than {\scriptsize EDGE\_CUT} (input parameter) are filtered out in Line 4. So only cluster points closer to the hypothesis boundary are chosen for training. Sorting in Line 6, before ordered matching in Line 9, adds the weighted aspect to the matching. It enables the heaviest edges to be picked earliest for contraction, subtly addressing both the main objectives of the directional aspect. Since the heaviest edges cover the classification patterns, prioritized selection of them results in uniform size of partitions. Prioritized selection also means that the most significant patterns are given preference, which conversely means the least significant patterns are avoided. So the hypothesis boundary, along which the least significant patterns reside, is orthogonal to the contracted edges, where the most significant patterns reside. Higher prediction accuracies are obtained because of the preference of contraction of the heaviest edges.

\begin{algorithm}
\label{algo:Algorithm3PartI}
\caption{\textbf{3, Part I :} Partial Weighted Matching}
\begin{flushleft}
Input : $G(V,E)$, \text{\scriptsize{EDGE\_CUT}}\\ 
Output : $M(G)/matching[]$ - matching of graph $G$
\end{flushleft}
\begin{algorithmic}[1]
\AlgphaseNotext{}
\State $edge\_list[] \gets \phi$
\ForAll {$i \,\epsilon \, V $}
  \ForAll {$j \,\epsilon \,neigh\_list(i) $}
    \If {$ w_{i,j}\, >= \,\text{\scriptsize{EDGE\_CUT}}$}
      \State $edge\_list \,{+}{=}\, [i,j]$
    \EndIf
  \EndFor
\EndFor
\State $sort(edge\_list)$ \algorithmiccomment{sort all collected edges}
\State $matching[] \gets \phi $
\State $visited[] \gets false $
\ForAll {$ [i,j] \text{ } \epsilon \,edge\_list $}
  \If {not $visited [i]$ and not $visited[j]$}
    \State $matching \,{+}{=}\, [i,j]$ \algorithmiccomment{add edge to matching}
    \State $visited[i] \,\gets \, true$
    \State $visited[j] \, \gets \, true$
  \EndIf
\EndFor
\end{algorithmic}
\end{algorithm}

% ===========================Algorithm 4====================
The coarsening formulation of Part~\hyperref[algo:Algorithm3PartII]{II} applies the directional aspect, as edge contraction occurs in this part. It is to be noted that the coarsening formulation is different in aim to otherwise researched formulations. Most of the popular formulations are intended for reducing communication cost or preserving the global structure while getting a low-cost representation of data \cite{InderjitSDhillon}. Furthermore, unlike Kernighan-Lin, and other matching based coarsening objectives, the presented optimization objective is deterministic in execution. 

In the part, re-assessment of target class, and edge list in Lines 6, and 10, respectively, for newly contracted nodes augments the standard coarsening step in Line 3. By using different values of $C_I$, and $C_E$ for initial versus re-assessment edge weights, the precedence of the kind of edges is established in the matching scheme. Original heavy edges are proiritized over re-assessed edges of newly contracted edges, achieving the orthogonality property. Transition of coarsening from original edges to newly contracted nodes can be captured by the virtue of drastic decrease in graph cost metric, presented in Eq.~\hyperref[eqn:GraphCost]{3},
\begin{eqnarray}
\text{cost}(G) = \sum_{e\epsilon E'} w_e,
\label{eqn:GraphCost}
\end{eqnarray}

where \ac{cg} is cost of the graph $G$, \ac{e} is an edge, \ac{Ep} denotes the subset of $E$ (set of all edges) such that $\{e \mid e \epsilon E \mbox{, and $>$ {\scriptsize{EDGE\_CUT}}}\}$, and \ac{we} is the weight of edge $e$.

One technical choice is to use {\scriptsize{MIN}$(i,j)$} in Line 4, for identifying the new node that is a result of the contraction of edge between nodes $i$, and $j$.

\begin{algorithm}
\label{algo:Algorithm3PartII}
\caption{\textbf{3, Part II :} Graph Coarsening (with edge weight reassessment)}
\begin{flushleft}
Input : $G(V,E), critical\_nodes[], matching[]$\\
Output : $G(V^\prime,E^\prime)$ - coarsened graph with $V^\prime\subseteq V, E^\prime\subseteq E, critical\_nodes[]$ 
\end{flushleft}
\begin{algorithmic}[1]
\AlgphaseNotext{}
\State $new\_critical\_nodes[] \gets \phi$
\ForAll {$[i,j]\, \epsilon \, matching$}
  \State $edge\_contraction(i,j)$\algorithmiccomment {standard graph coarsening step}
  \State $critical\_nodes \,{+}{=} \,\text{\scriptsize{MIN}}(i,j)$\algorithmiccomment{add new critical node}
  \State $new\_critical\_nodes\,{+}{=}\,\text{\scriptsize{MIN}}(i,j)$
  \State $update\_tc(\text{\scriptsize{MIN}}(i,j))$ \algorithmiccomment {re-evaluating target class of new critical node}
\EndFor
\ForAll {$i\,\epsilon \,new\_critical\_nodes$}
  \ForAll {$j\,\epsilon \,neigh\_list(i)$}
    \If {$j\,\epsilon \,critical\_nodes$}
      \State $update\_w(i,j)$ \algorithmiccomment{re-assess edge weights}
    \EndIf
  \EndFor
\EndFor
\end{algorithmic}
\end{algorithm}

Part I, and II, are executed in the iterations of Part~\hyperref[algo:Algorithm3PartIII]{III} in Lines 5, and 6, respectively. In Line 7, kink detection in graph cost is used as a termination condition of the iterative algorithm. However, a maximum number of coarsening iterations, i.e. {\scriptsize MAX\_NUM\_OF\_COARSENING\_ITER} in Line 4, is used in the majority of tests. This step concludes the formulation of the heuristic. It is referred to as \ac{gch} for the remainder of this work. Similarly, edge cut is referred as GC edge cut.

\begin{algorithm}
\label{algo:Algorithm3PartIII}
\caption{\textbf{3, Part III :} Graph Clubbing}
\begin{flushleft}
Input : $G(V,E)$, \text{\scriptsize{EDGE\_CUT}} \\ 
Output : $G(V',E')$ - iteratively coarsened graph
\end{flushleft}
\begin{algorithmic}[1]
\AlgphaseNotext{}
\State $prev\_gc \gets \text{cost}(G)$ \algorithmiccomment{graph cost equation, Eq. 3}
\State $iteration \gets 0$
\State $critical\_nodes[] \gets \phi$

\While {$iteration < \text{\scriptsize{MAX\_NUM\_OF\_COARSENING\_ITER}}$}
  \State $matching \gets \text{PWM}(G, \text{\scriptsize{EDGE\_CUT}})$ \algorithmiccomment{Part I}
  \State $G, critical\_nodes \gets \text{GC}(G, critical\_nodes, matching)$ \algorithmiccomment{Part II}
  \State $curr\_gc \gets \text{cost}(G)$ \algorithmiccomment{current graph cost}
  \If {$kink\_detected (curr\_gc,prev\_gc)$} \algorithmiccomment{backward difference gradient}
    \State $break$
  \Else 
  \State $prev\_gc  \gets curr\_gc$ \algorithmiccomment{march forward}
  \State $iteration++$
  \EndIf
\EndWhile
\end{algorithmic}
\end{algorithm}

The implementation of few optimizations improved the run-time. First, the starting nodes in Line 2, Part~\hyperref[algo:Algorithm3PartI]{I}, are the ones that are identified relevant in GSH. After each iteration, half of the nodes that belong to contracted edges are reduced in the update of relevant nodes list ($ relevant\_nodes\_list$ in Algorithm 2). As a result, the complexity of the matching algorithm reduces with the coarsening iterations. Second, the neighbor list of a node is sorted. As a result, edge contraction computation is linear (in complexity) to $nn$. It is computationally canonical to the sorted union of two lists. Lastly, usual numerical optimizations, such as masking to avoid dynamic memory allocations, and indexing (at the expense of memory) for $O(1)$ searching are used.

\subsubsection*{Network design}
An event-driven, multi-process algorithm is designed to distribute the partitions obtained after GCH in a communication-free manner. In the network application, processes assume a (single) master or (multiple) worker role. Part~\hyperref[algo:Algorithm4PartI]{I}, and~\hyperref[algo:Algorithm4PartII]{II} of Algorithm~\hyperref[algo:Algorithm4PartI]{4}, respectively describe master's, and worker's side of event-handling design. 

For the master process which executes Part~\hyperref[algo:Algorithm4PartI]{I}, one partition from the list of partitions, $parts\_list[]$, is communicated to the requesting worker process (the one which issues {\scriptsize{DATA\_REQUEST}} event in Line 3 of Part~\hyperref[algo:Algorithm4PartII]{II}) by trigger of {\scriptsize{DATA}} event in Line 6. The worker process, upon receiving the event {\scriptsize{DATA}}, proceeds to training with the recieved data partition in Line 5 of Part II. After collecting acknowledgements of the completion of training of all data partitions in Line 9, the master process terminates every worker by issuing {\scriptsize{TERM\_TRAIN}} event. The design implements a round-robin scheme, which balances load of the network queries. That is accompolished by having a queue data structure for recording {\scriptsize{DATA\_REQUEST}} events in Line 3. It is to be noted that conflict of two simultaneous entries is resolved by time stamps, making the queue fair with respect to a worker's request.

\begin{algorithm}
\caption{\textbf{4, Part I :} Master Process}
\label{algo:Algorithm4PartI}
\begin{flushleft}
Input : $num\_dat\_parts, parts\_list[]$ \\
Output : $num\_dat\_parts$ number of trained hypotheses
\end{flushleft}
\begin{algorithmic}[1]
\AlgphaseNotext{}
\State $curr\_dat\_part\_no \gets -1 $
\State $train\_ackn\_no \gets 0 $
\Event {\text{\scriptsize{DATA\_REQUEST}}}
  \If {$curr\_dat\_part\_no < num\_dat\_parts $}
    \State $curr\_dat\_part\_no++$
    \GenEvent{\text{\scriptsize{DATA}}} \algorithmiccomment {sending a part to the requesting worker}
    \EndGenEvent
  \EndIf
\EndEvent

\Event{\text{\scriptsize{DONE\_TRAINING}}}
  \State $train\_ackn\_no++$
  \If {$train\_ackn\_no = num\_dat\_parts $}
    \ForAll {$workers$}
      \GenEvent{\text{\scriptsize{TERM\_TRAIN}}}
      \EndGenEvent
    \EndFor
  \EndIf
\EndEvent
\end{algorithmic}
\end{algorithm}

\begin{algorithm}[H]
\caption{\textbf{4, Part II :} Worker Process}
\label{algo:Algorithm4PartII}
\begin{algorithmic}[1]
\State $term\_train \gets \text{\scriptsize{FALSE}}$
\While{not $term\_train$}
  
  \GenEvent{\text{\scriptsize{DATA\_REQUEST}}}
  \EndGenEvent

  \Event{\text{\scriptsize{DATA}}}
    \State $training$ \algorithmiccomment {training of received part}
  \EndEvent

  \GenEvent{\text{\scriptsize{DONE\_TRAINING}}}
  \EndGenEvent  

  \Event{\text{\scriptsize{TERM\_TRAIN}}}
    \State $term\_train = \text{\scriptsize{TRUE}}$
  \EndEvent
\EndWhile
\end{algorithmic}
\end{algorithm}

\begin{figure}[h!]
      \centering \includegraphics[width=\columnwidth]{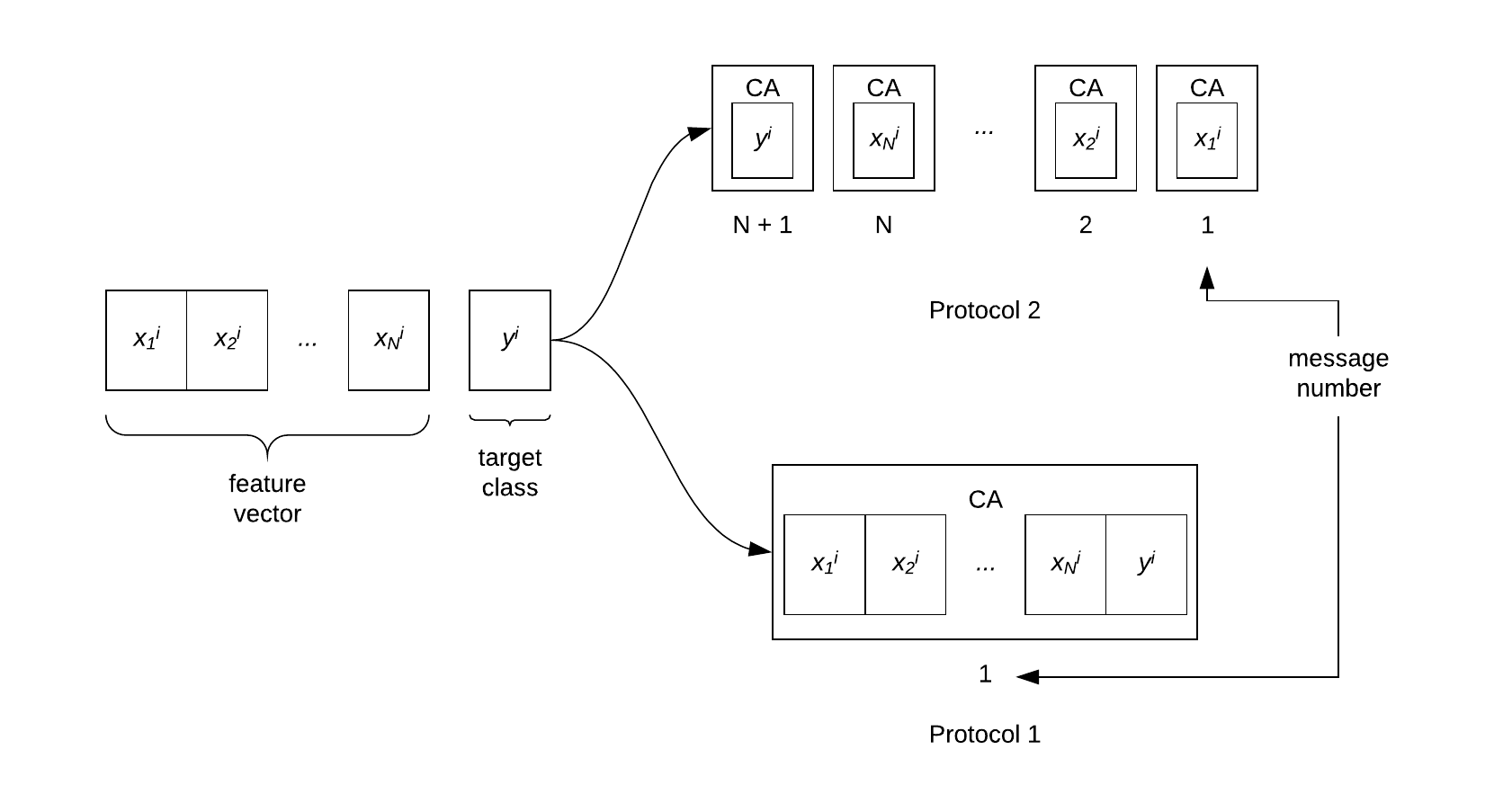}
      \caption{\csentence{Messaging protocols.} Protocol 2, i.e. single entry versus Protocol 1, i.e. marshaling protocol is presented for data of the \ensuremath{i^{\text{th}}} point.}
      \label{fig:1}
\end{figure}

Instead of directly using TCP channels for message communication, the distributed messaging API ZeroMQ \cite{zmq} is used. It provides essential safety, and liveness properties on network channels. However, apart from message guarantees such as the liveness, and safety property of ‘once only message delivery’, ZeroMQ is rudimentary compared to higher level message passing libraries. This gives an opportunity to design, and optimize various aspects of the architecture. One such aspect is the messaging protocol. A couple of messaging protocols are designed as shown in Figure~\hyperref[fig:1]{1}. Protocol 2 implementes a single float/double entry (in character array or CA) messaging scheme, whereas protocol 1 first requires marshaling all entries of a data point before messaging it.

Another aspect that was tested is connection time of the network. Connection time is measured on the master process, and included the following steps:
\begin{itemize}
\item Start of TCP channels (wrapped in the API)
\item Initialize a hash table
\item Recieve connection request from all worker processes
\item Send connection confirmation to all worker processes
\end{itemize}

The connection prodecure requires step 2 for maintaining worker processes' information, giving an opportunity to optimize the step as per the need. A light-weight hash table ($\approx 0.6 MB$), and hash key is designed which generates unique keys for $< 100$ worker processes. The design helps to reduce the overhead of starting, and running the multi-process application.

\subsection*{Pre-processor for the testing phase}
Unlike the application of GSH, which results in a single training set, few data partitions are obtained after GCH, and the training set is the union of these partitions. It means that there would be as many classifiers as the number of partitions. Hence, it is needed to determine which classifier to choose for predicting a point from the testing dataset, $V_{T}$. Algorithm~\hyperref[algo:Algorithm5]{5}, nearest hypothesis search (\ac{nhs}), is used for this task. A search space is formed consisting of nodes of the coarsened graph in Line 1. ANN search for the nearest hypothesis follows in Line 5.

\begin{algorithm}

\caption{\textbf{5 :} Nearest Hypothesis Search}
\label{algo:Algorithm5}
\begin{flushleft}
Input : $V_{T}$ - testing dataset, $G(V, E)$ - graph after GCH \\ 
Output : $nearest\_hypothesis[]$ - nearest hypothesis for every point in $V_{T}$
\end{flushleft}
\begin{algorithmic}[1]
\AlgphaseNotext{}
\State $search\_space\gets V$
\AlgphaseNotext{}
\State $nearest\_hypothesis[]\gets \phi$
\State $flann() \gets search\_space$ \algorithmiccomment {FLANN instance}
\State $flann.index()$
\State $nearest\_hypothesis \gets flann.ann\_search(1, V_{T})$\algorithmiccomment {only nearest hypothesis}
\end{algorithmic}
\end{algorithm}

Once the nearest hypothesis is determined, prediction of the target class for testing data points follows. This added step before the testing phase only takes about 5~-~7\% of the run-time of the testing phase for SVM class of algorithms, as will be shown in the `Results' section.

\section*{Results}
Results are presented in two major divisions, first with tests on parameter space of the heuristic, and second for gauging performance of the heuristic. All the tests were conducted on a variety of datasets.

\subsection*{Parameter space of heuristic}
\begin{figure}[h!]
      \centering \includegraphics[width=0.5\columnwidth]{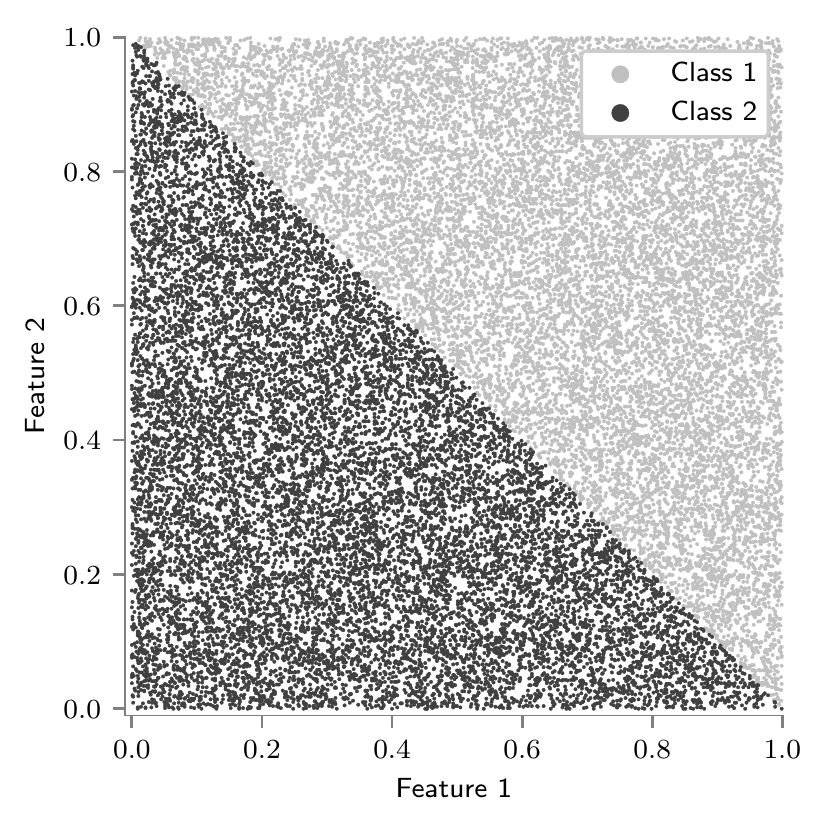}
      \caption{\csentence{Near linearly-separable dataset.}}
      \label{fig:2}
\end{figure}

In this set of tests, the focus is on working details, and exemplifying steps of the heuristic. Datasets similar to that shown in Figure~\hyperref[fig:2]{2}, with parameters summarized in Table~\hyperref[tab:table_1]{1} are extensively used.

\begin{table}[h!]
\caption{Dataset parameters.}
  \begin{tabular}[b]{|c|c|}\hline
  Parameter & Value \\
  \hline
  $n$ & 30000 \\
  \hline
  $d$ & 2 \\
  \hline
  $n_{c}$ & 300 \\
  \hline
  $C_I$ & $e^{1.0}$ \\
  \hline
  $C_E$ & $e^{4.0}$ \\
  \hline
  GS edge cut & $3.01$\\
  \hline
  GC edge cut & $3.20$\\
  \hline
  \end{tabular}
  \label{tab:table_1}
\end{table}

\subsubsection*{Node reach, and ENS}
Tests in this section present heuristic tools that capture original classification data into the weighted graph. These tools are designed to handle real datasets, which vary diversely in characteristics. A mix of real, and synthetic datasets that mimic varying characteristics are considered.

\begin{figure}[h!]
    \centering \includegraphics[width=\columnwidth]{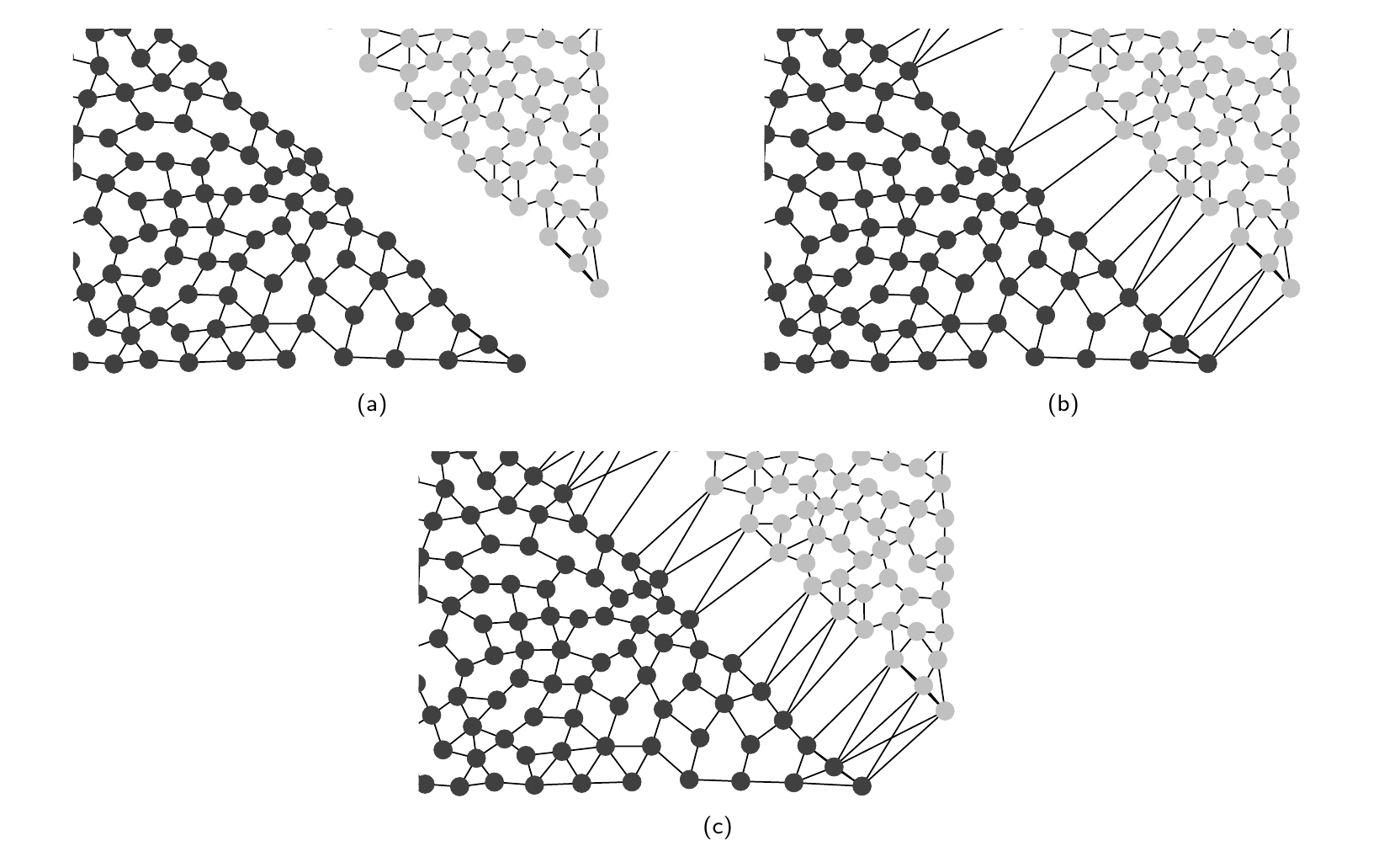}
      \caption{\csentence{Exclusive neighbor search.} Number of ENS iterations are 0, 2, and 4 in (a), (b), and (c), respectively. Zoomed view of dataset in Figure 2.}
      \label{fig:3}
\end{figure}

A timeline of the ENS procedure with $nn = 4$ is presented in Figure~\hyperref[fig:3]{3}. It is based on the dataset of Table~\hyperref[tab:table_1]{1}. However, class 2 data points were intentionally translated to create separation, which is very typical for real data. It is evident that the connectivity of the graph increases with more iterations. Skewness control of the constructed graph, explained in Part III of Algorithm 1, was carried out at end of the iterations of the ENS procedure, resulting in reduction of available nodes for search as shown in Table~\hyperref[tab:table_2]{2}.

\begin{table}[h!]
\caption{Search space reduction.}
  \begin{tabular}[b]{|c|c|c|}\hline
    \# iterations & \# class 1 nodes & \# class 2 nodes \\ \hline
    0 & 143 & 157\\
    \hline
    2 & 143 & 157\\
    \hline
    4 & 139 & 153\\
    \hline
  \end{tabular}
  \label{tab:table_2}
\end{table}

\begin{figure}[h!]
    \centering \includegraphics[width=\columnwidth]{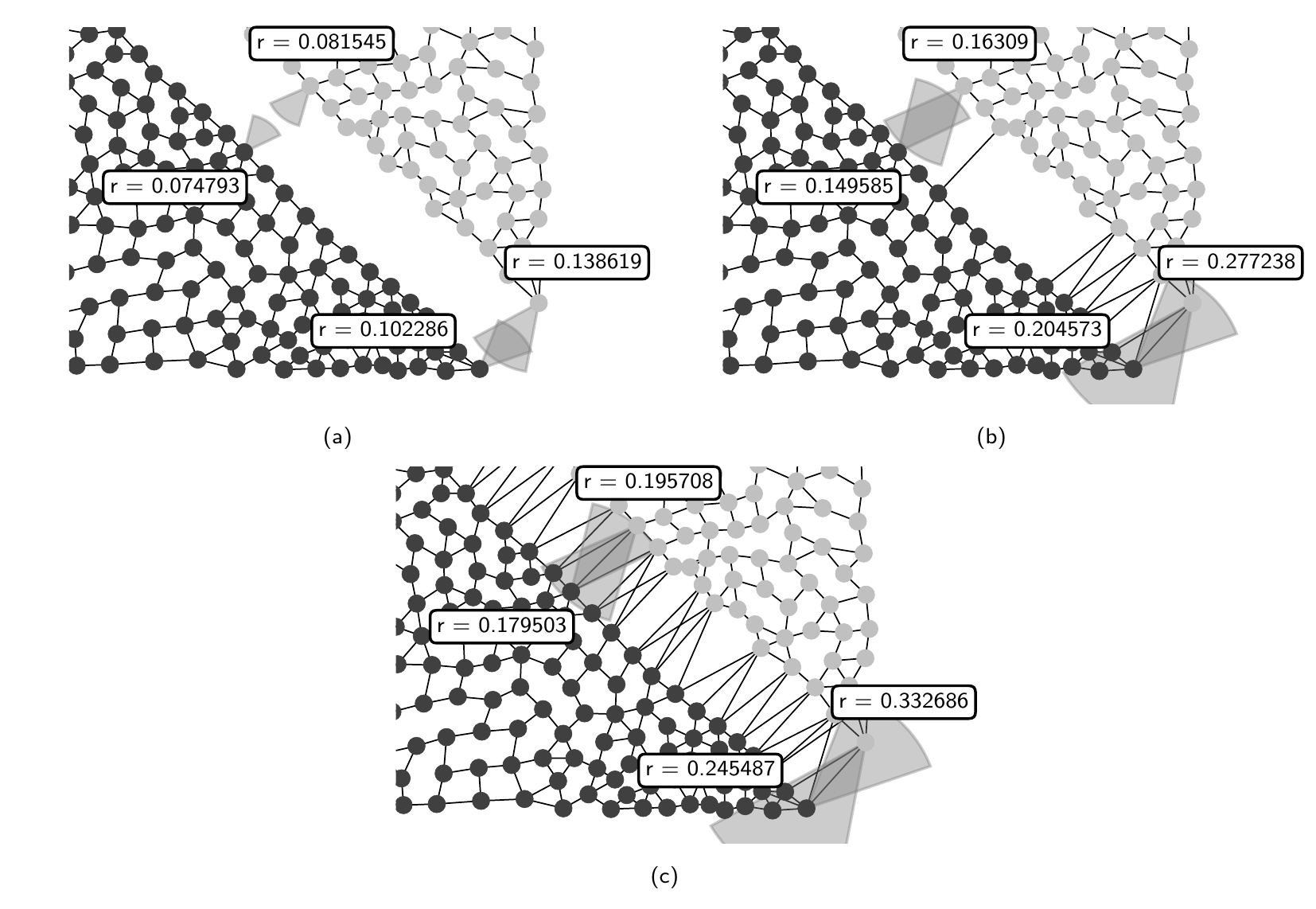}
      \caption{\csentence{Reach controlled graph knitting.} Scaling constant $R$ is $0.5$, $1.0$, and $1.25$ in (a), (b), and (c), respectively. Zoomed view of dataset in Figure 2.}
      \label{fig:4}
\end{figure}

A second way to control connectivity is fine tuning of the reach equation, presented in Eq.~\hyperref[eqn:reach]{1}. In the next test, scaling constant $R$ is varied, and results are presented in Figure~\hyperref[fig:4]{4}. Three cases are shown in sucession, under-reach (a), ideal-reach (b), and over-reach (c). Even in the over-reach case, inner nodes are not able to make opposite class neighbors, enforcing planar construction of the graph.

\begin{figure}[h!]
    \centering \includegraphics[width=\columnwidth]{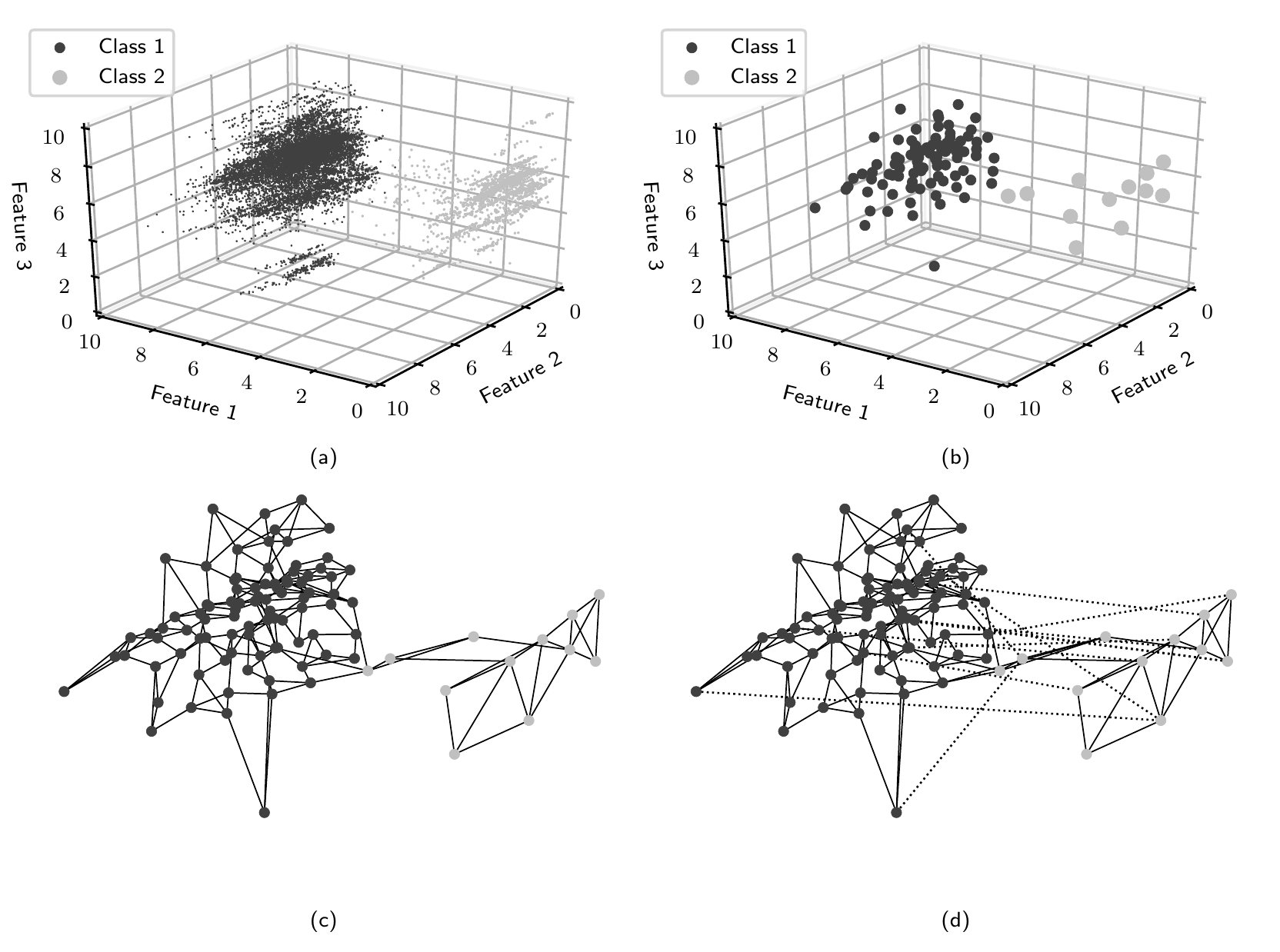}
      \caption{\csentence{Graph knitting procedure.} Scaled UCI dataset, clustered data, constructed graph with no, and 1 ENS iteration in (a), (b), (c), and (d), respectively. Dashed edges in (d) are formed after ENS procedure.}
      \label{fig:5}
\end{figure}

The ENS procedure was next run on a real dataset, called cuff-less blood pressure estimation dataset from UCI machine learning repository \cite{UCI_2, UCI}. It is a three attribute, 12000 instances, real, multivariate dataset. Results are presented in Figure~\hyperref[fig:5]{5}, with Subfigure (a) showing the dataset, (b) showing cluster centers, whereas (c), and (d) showing the constructed graph without, and with ENS procedure, respectively. The latter graph is connected because necessary edges are present between opposite class nodes, covering the entire hypothesis. Class imbalance is analysed next, and results are presented in Table~\hyperref[tab:table_3]{3}. The column 2 shows the number of data points for each class originally, and column 3 shows the number of data points after ENS, as presented in the graph of Figure~\hyperref[fig:5]{5}(d). Imbalance of target classes, quantified by standard deviation (\ac{sd}) is significantly reduced after ENS.

\begin{table}[h!]
\caption{Class imbalance.}
  \begin{tabular}[b]{|c|c|c|c|}\hline
    Procedure & \# class 1 points & \# class 2 points & SD \\ \hline
    None & 1743 & 10256 &  4256.5\\
    \hline
    GSH & 1661 & 2001 & 170.0\\
    \hline
  \end{tabular}
  \label{tab:table_3}
\end{table}

\subsubsection*{GSH}
\begin{figure}[h!]
    \centering \includegraphics[width=\columnwidth]{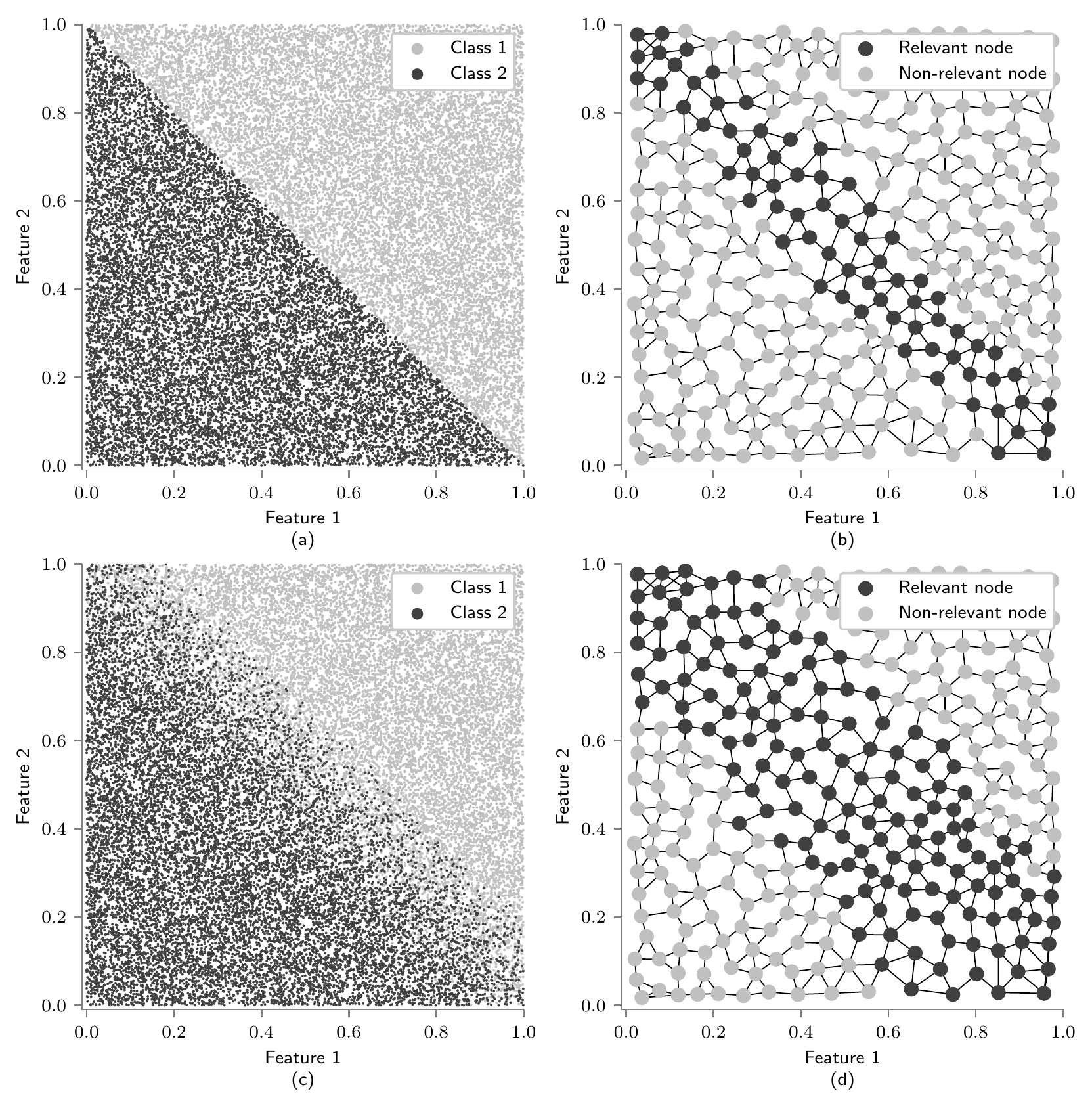}
      \caption{\csentence{Reduction of training set.} Classification data is more confused in (c) compared to (a). (b), and (d) show cluster nodes that were selected for (a), and (c) cases, respectively.}
      \label{fig:6}
\end{figure}

The aim of this test is to exemplify the two-fold edge weight scheme, shown in Eq.~\hyperref[eqn:PowerScheme]{2}. It shows the role of edge weights in the outcome of GSH, summarized in Figure~\hyperref[fig:6]{6}. As classification patterns in the underlying data gets more confused in succession in Subfigures (a), and (c), more edges are weighted significantly, resulting in the selection of a bigger training set by GSH.

\subsubsection*{GCH}
This test exemplifies the formulation of the graph clubbing scheme. Although the edge weight scheme is important in the application of GSH, it is primarily designed to play a crucial role in the priority/directional aspect of the coarsening objective. Constants for initial edge weights were such that $C_E$ was kept significantly higher than $C_I$, as shown in Table~\hyperref[tab:table_4]{4}. Two cases of re-assessments were considered, namely case I, and case II.

\begin{table}[h!]
\caption{Pattern measure constants.}
  \begin{tabular}{|c|c|c|c|}
  \hline
  \multirow{2}{*}{Parameter} & \multirow{2}{*}{Initial} & \multicolumn{2}{c|}{Re-assessment}\\
  \cline{3-4}
  & & Case I & Case II\\
  \hline
  $C_I$ & $e$ & $e^{1.5}$ & $1$\\
  \hline
  $C_E$ & $e^5$ & $1$ & $1$\\
  \hline
  \end{tabular}
  \label{tab:table_4}
\end{table}

\begin{figure}[h!]
    \centering \includegraphics[width=\columnwidth]{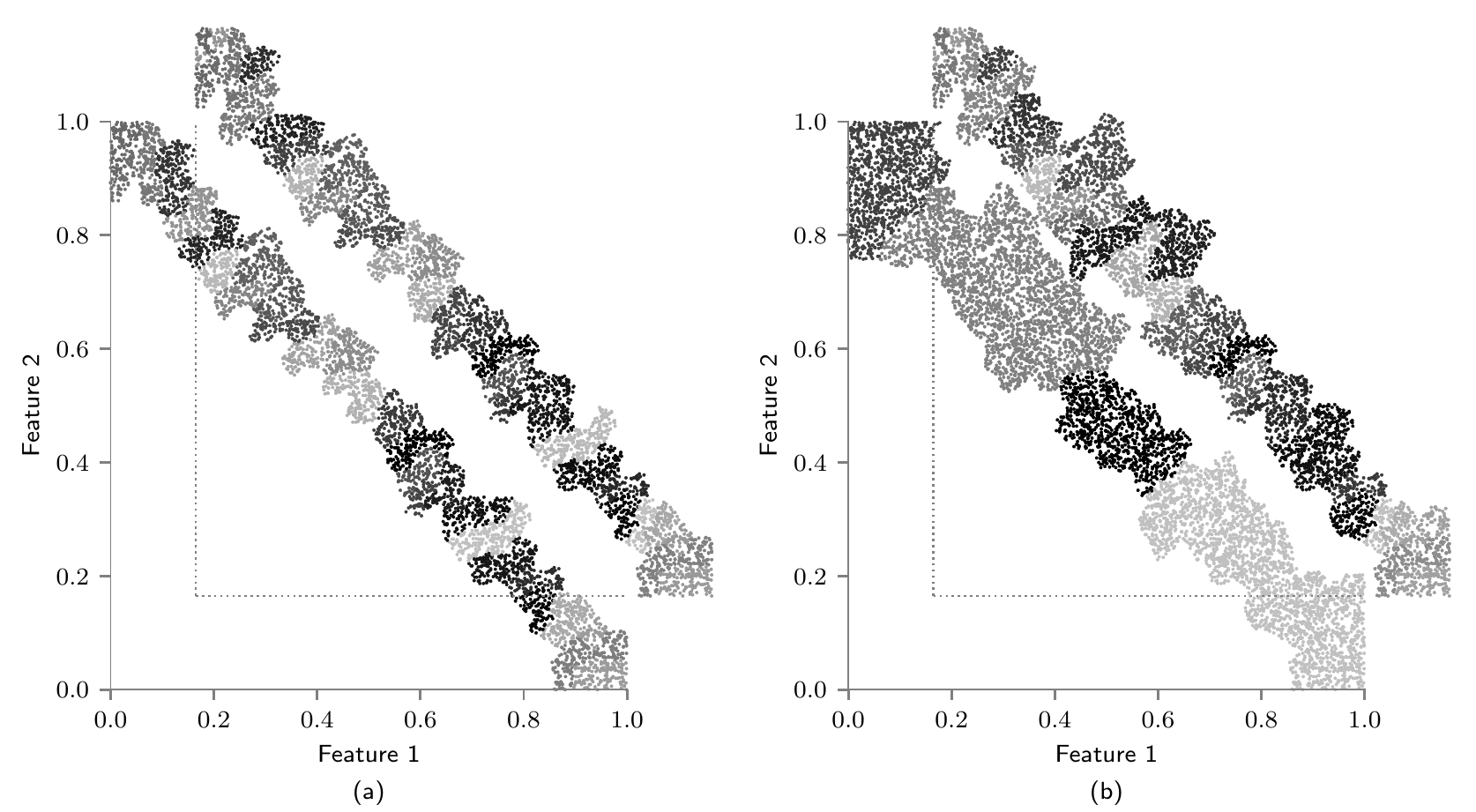}
      \caption{\csentence{Partitioning of training set.} Number of coarsening iterations is 2, and 10 for (a), and (b), respectively. In each Subfigure, the first plot is for the re-assessment case I, whereas the overlaying plot with shifted origin is for the case II.}
      \label{fig:7}
\end{figure}

\begin{figure}[h!]
    \centering \includegraphics[width=0.5\columnwidth]{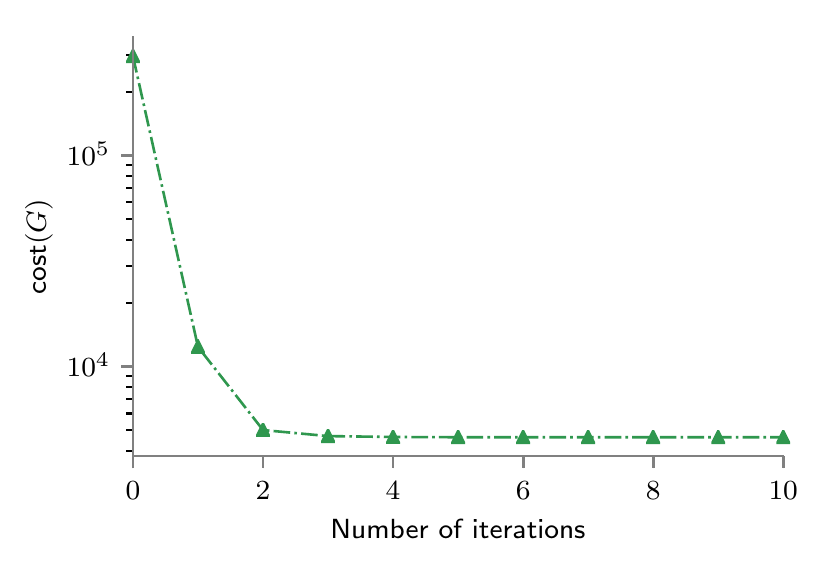}
      \caption{\csentence{Graph cost function with coarsening iterations.}}
      \label{fig:8}
\end{figure}

Since initial constants were kept significantly higher than their re-assessment counterparts, original nodes were contracted first in both cases, as can be seen in Figure~\hyperref[fig:7]{7}(a), and (b). Transition to contraction of re-assessed type nodes is reflected in the graph cost characteristic in Figure~\hyperref[fig:8]{8} for case I. After the \ensuremath{2^{\text{nd}}} iteration, the slope magnitude decreases significantly because of contraction of all original nodes, that have significantly higher edge weights dictated by heavier initial constants. Edge contraction continues to aggressively club the graph for case I, including the re-assessed nodes. This results in fewer partitions compared to case II in Figure~\hyperref[fig:7]{7}(b). Clubbing is almost shut off in case II because of trivial re-assessment constants. Note that in general, the more the coarsening iterations, the fewer the partitions.

\subsection*{Performance evaluations}
Several tests were conducted to evaluate performance of the presented methods, including GSH, serial GCH, and distributed GCH. In addition, the network architecture for distributed GCH is evaluated.

Dataset I was a synthetically constructed two-dimensional, near-linearly separable dataset with parameters shown in Table~\hyperref[tab:table_5]{5}. Dataset II was a similar dataset, but in place of a (nearly) straight-separating hyperplane, a spherical separating hyperplane of radius 0.2 was employed. Otherwise, it uses the same parameters as Dataset I. A dataset from the UCI machine learning repository, called skin segmentation dataset \cite{UCI, UCI_1} is the next dataset, called Dataset III for the remainder. This classification dataset has four numeric features, and $\approx$245k observation instances. The nominal VC dimension is $245$. Table~\hyperref[tab:table_6]{6} summarizes other relevant parameters.

\begin{table}[h!]
\caption{Dataset I \& II parameters.}
  \begin{tabular}{|c|c|c|}
    \hline
    \multicolumn{2}{|c|}{ Parameter} & Values\\
    \hline
    \multirow{2}{*}{Range of \# data points} & testing & 3k~-~300k\\
    \cline{2-3}
    & training & 1k~-~100k\\
    \hline
    \multirow{3}{*}{\# iterations} & coarsening & 10\\
    \cline{2-3}
    & clustering & 3~-~5\\
    \cline{2-3}
    & ENS & 0\\
    \hline
    \multicolumn{2}{|c|}{$nn$} & 4\\
    \hline
    \multicolumn{2}{|c|}{$d$} & 3\\
    \hline
    \multicolumn{2}{|c|}{Range of $n_{c}$} & 10~-~1000\\
    \hline
    \multicolumn{2}{|c|}{ $R$ } & 1.0~-~2.0\\
    \hline
  \end{tabular}
  \label{tab:table_5}
\end{table}

\begin{table}[h!]
\caption{UCI dataset, called Dataset III parameters.}
\begin{tabular}{|c|c|c|}
  \hline
  \multicolumn{2}{|c|}{ Parameter} & Values\\
  \hline
  \multirow{2}{*}{\# data points} & test & 122529\\
  \cline{2-3}
  & training & 122528\\
  \hline
  \multirow{3}{*}{\# iterations} & coarsening & 0\\
  \cline{2-3}
  & clustering & 5\\
  \cline{2-3}
  & ENS & 0\\
  \hline
  \multicolumn{2}{|c|}{$nn$} & 5\\
  \hline
  \multicolumn{2}{|c|}{$d$} & 4\\
  \hline
  \multicolumn{2}{|c|}{$n_{c}$} & 500\\
  \hline
  \multicolumn{2}{|c|}{ $R$ } & 1.0~-~2.0\\
  \hline
\end{tabular}
\label{tab:table_6}
\end{table}

Note that the computation setup was kept consistent. Programming language of all codes is C/C++ for which computation time was measured. That includes every step of the presented heuristic, and the classification algorithms. Lastly, time stamping was carried on an otherwise idle system. 

\subsubsection*{GSH}
The main aim of the next set of tests is to compare the training phase using GSH against state-of-the-art shrinking heuristic of LIBSVM. Classification algorithms used in these tests were all variants of LIBSVM's SMO implementation \cite{RongEnFan}, which is a method of the second type.

Two testing variables are evaluated. First, computation run-time of GSH along with the classification algorithm on reduced training data is compared against that of the same classification algorithm but augmented with the shrinking heuristic. Second, the prediction accuracy of the learned models obtained from both the above setups are compared.

\begin{figure}[h!]
    \centering \includegraphics[width=\columnwidth]{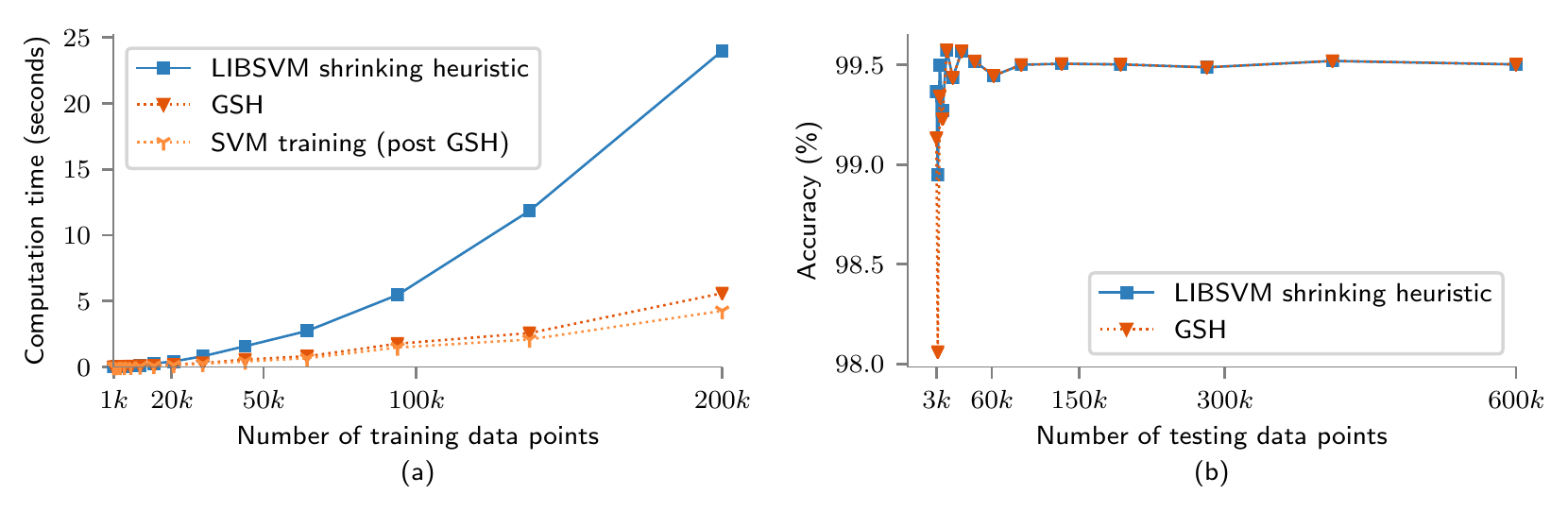}
      \caption{\csentence{C-SVM with linear kernel.}}
      \label{fig:9}
\end{figure}

Results are presented in the following way. Each of Figures~\hyperref[fig:9]{9}~-~\hyperref[fig:15]{15} is for a SVM classifier, that includes C-SVM, and nu-SVM with linear, polynomial, and rbf kernels. Subfigure (a) is used to present computation run-time results, and (b) is used for prediction accuracy comparison. Training dataset size was geometrically varied from 1k to 200k, and testing dataset size was kept three times that of the training. Accordingly, the parameter $n_{c}$ was varied geometrically in the range given in Table~\hyperref[tab:table_5]{5}. Note that training time is included in `LIBSVM shrinking heuristic', and `GSH' plots in Subfigure (a), but is not explicitly written for convenience. A third line-plot denoted `SVM training (post GSH)' is presented in Subfigure (a) to separate computation run-time of GSH from subsequent training.

C-SVM with linear kernel was the first classification algorithm tested, as presented in Figure~\hyperref[fig:9]{9}. The GSH scales even for a low number of points ($\approx$10k), as shown in Subfigure (a), however, scaling becomes more evident with more training data points. Furthermore, it is visible that GSH took a small fraction of total time, as indicated by the separation between the second, and third line plots. It highlights the scalability issue of the SVM formulation. In Subfigure (b), the prediction accuracy of reduced training closely follows that of full training, clearly visible after $\approx$20k testing data points.

\begin{figure}[h!]
    \centering \includegraphics[width=\columnwidth]{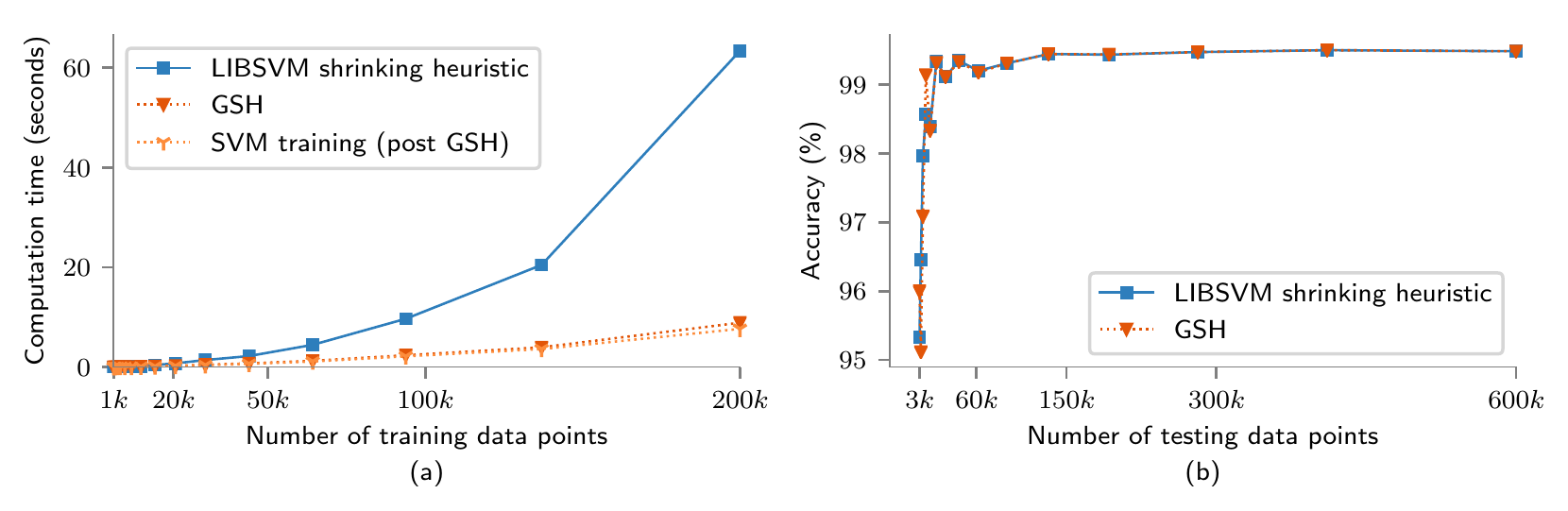}
      \caption{\csentence{C-SVM with polynomial kernel.}}
      \label{fig:10}
\end{figure}

C-SVM with polynomial kernel was the second classifier tested, as presented in Figure~\hyperref[fig:10]{10}. Observations here follow that of the earlier case. Although, one noticeable difference is a better run-time profile to the previous case. This is because of the higher run-time complexity of the polynomial kernel classifier than that of the linear kernel presented in Figure~\hyperref[fig:9]{9}.

\begin{figure}[h!]
    \centering \includegraphics[width=\columnwidth]{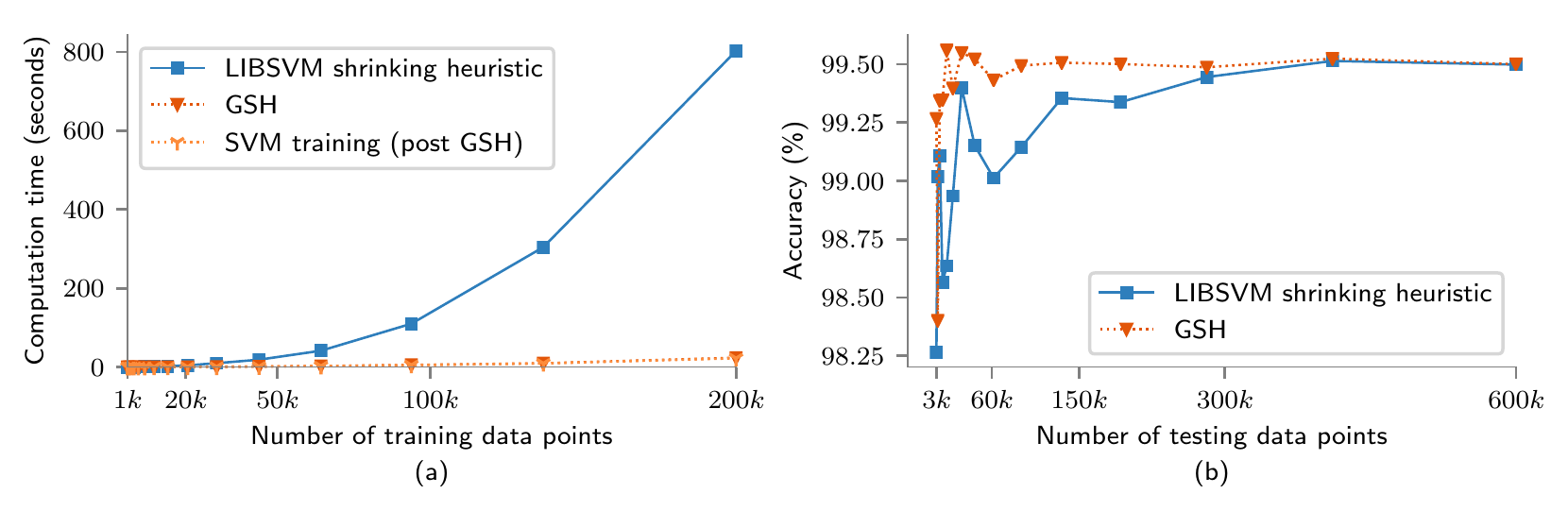}
      \caption{\csentence{nu-SVM with linear kernel.}}
      \label{fig:11}
\end{figure}

\begin{figure}[h!]
    \centering \includegraphics[width=\columnwidth]{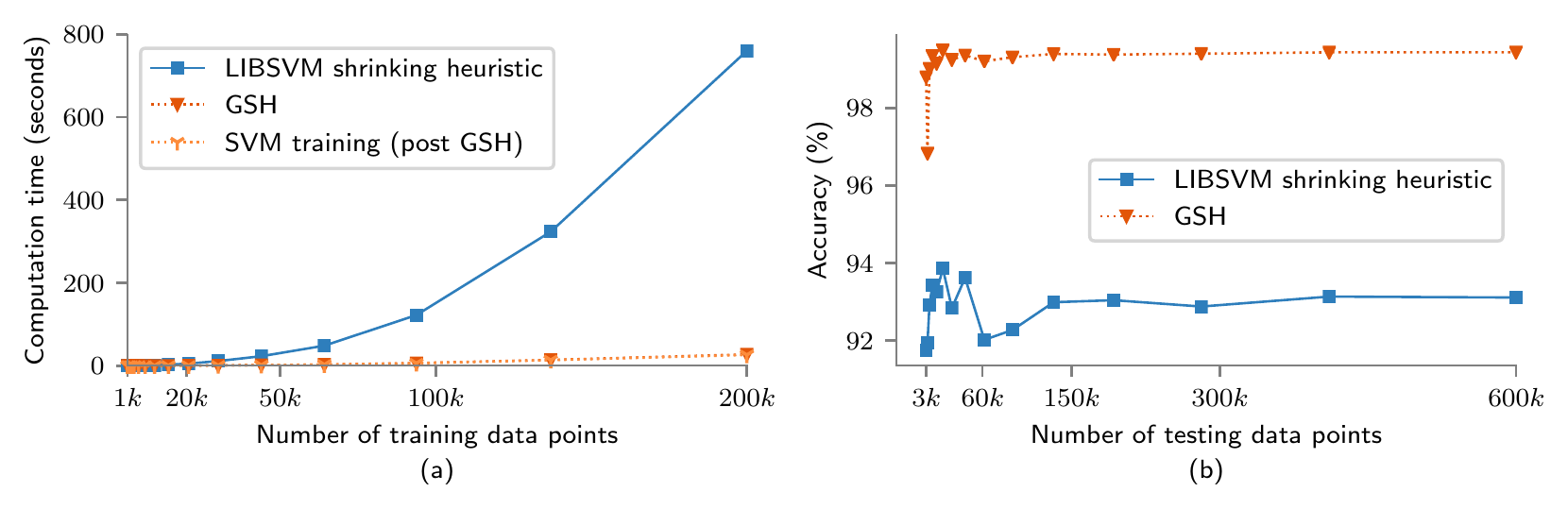}
      \caption{\csentence{nu-SVM with polynomial kernel.}}
      \label{fig:12}
\end{figure}

For the second SMO implementation, nu-SVM, Figures~\hyperref[fig:11]{11}, and~\hyperref[fig:12]{12} summarize linear, and polynomial kernel cases, respectively. Note that the training phase with nu-SVM in Figures 11(a), and 12(a) is significantly slower compared to C-SVM in Figures~\hyperref[fig:9]{9}(a), and \hyperref[fig:10]{10}(a). Furthermore, it is observed that reduced training can result in improved prediction accuracy, as can be seen in Figure~\hyperref[fig:12]{12}(b). For the polynomial kernel, GSH performed better than native training by  about 6\%.

\begin{figure}[h!]
    \centering \includegraphics[width=\columnwidth]{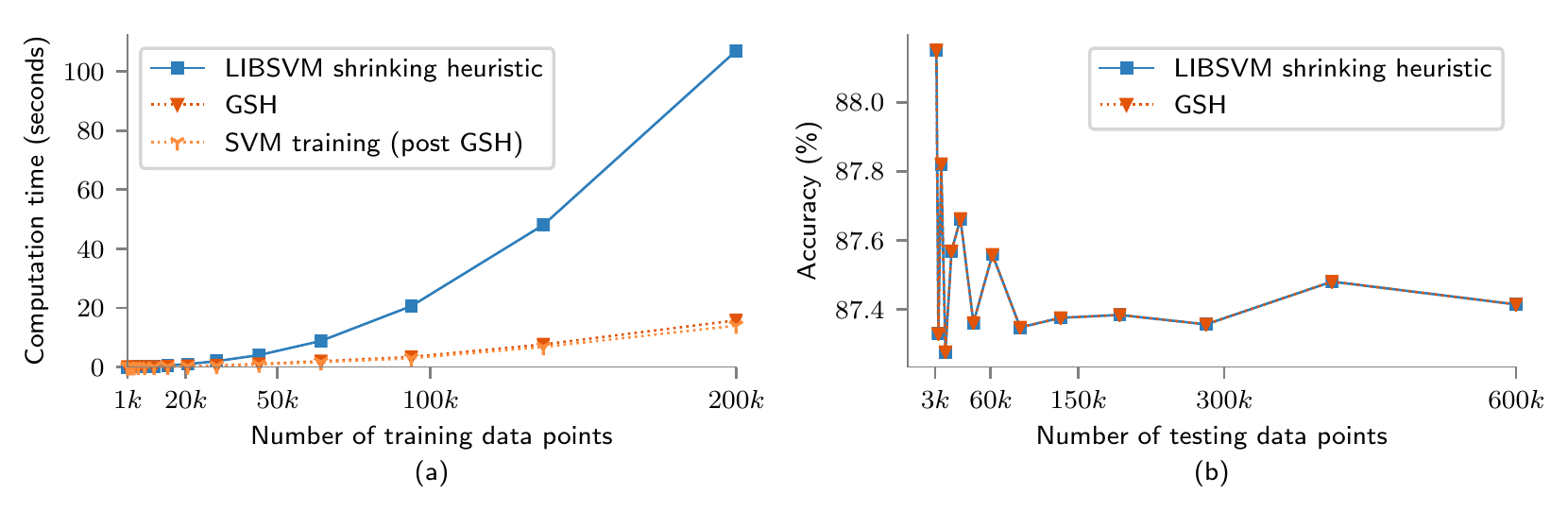}
      \caption{\csentence{C-SVM with linear kernel.}}
      \label{fig:13}
\end{figure}

Dataset I was used in all the tests until now, and near-perfect accuracy plots support that it is an ideal dataset. A more realistic case, Dataset II, was considered for Figures~\hyperref[fig:13]{13}~-~\hyperref[fig:15]{15}.

\begin{figure}[h!]
    \centering \includegraphics[width=\columnwidth]{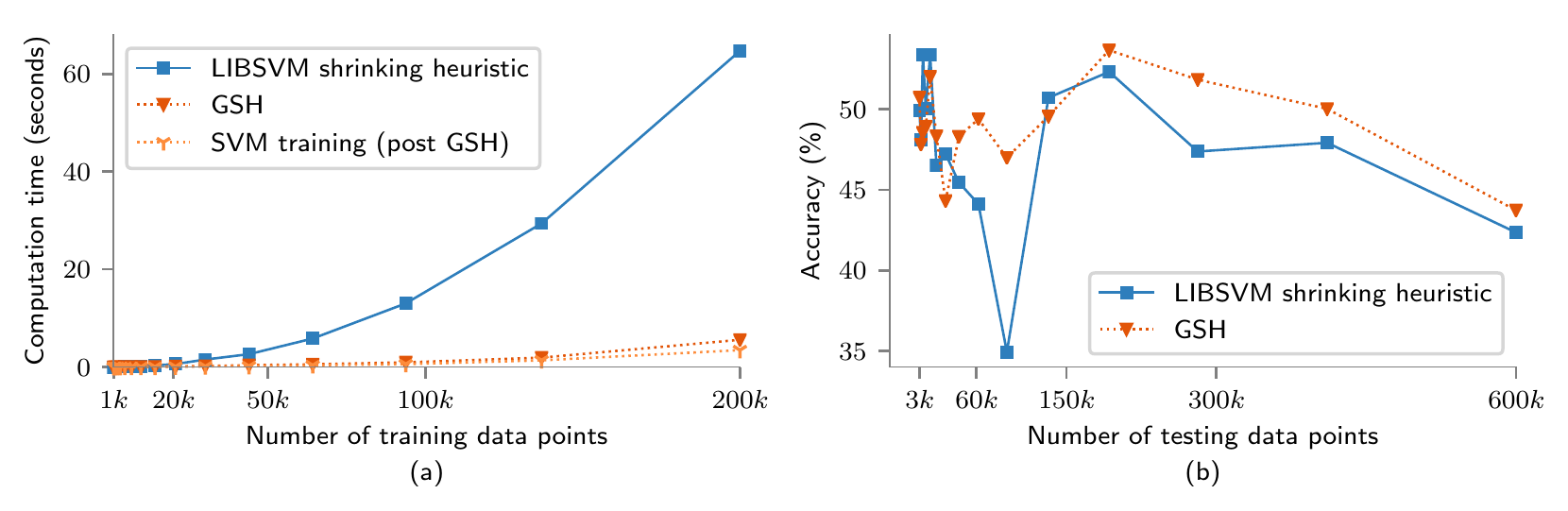}
      \caption{\csentence{nu-SVM with linear kernel.}}
      \label{fig:14}
\end{figure}

\begin{figure}[h!]
    \centering \includegraphics[width=\columnwidth]{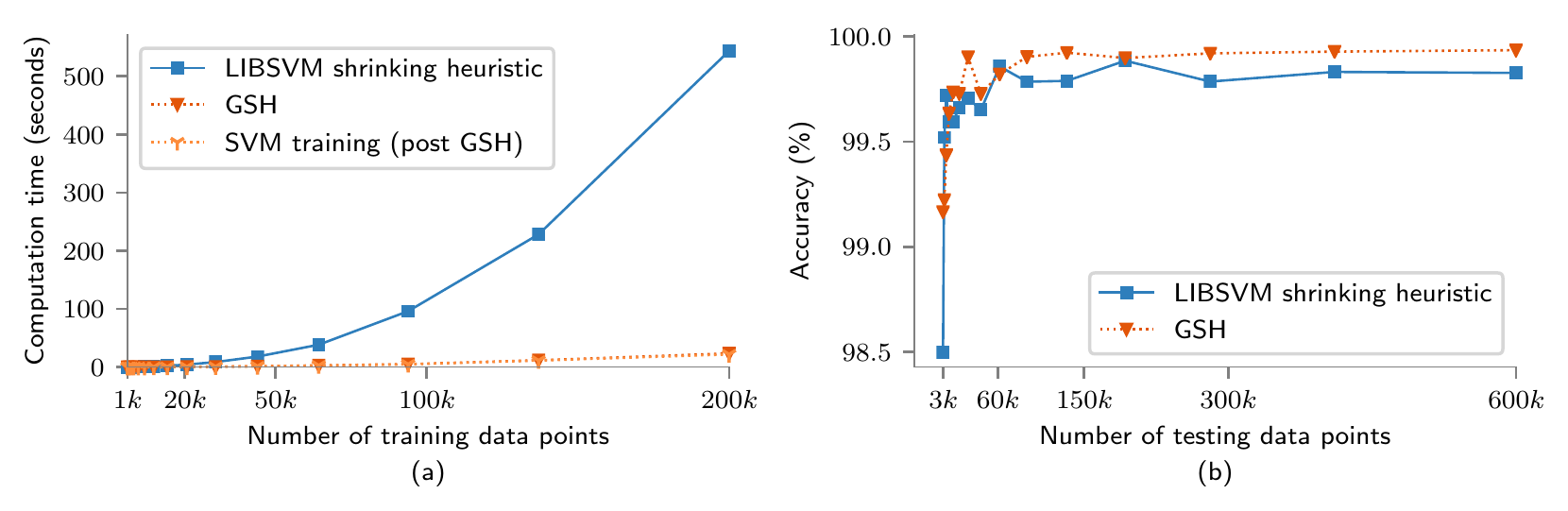}
      \caption{\csentence{nu-SVM with rbf kernel.}}
      \label{fig:15}
\end{figure}

Again, C-SVM with linear kernel was the first classification algorithm tested, as presented in Figure~\hyperref[fig:13]{13}. However, prediction accuracies are in a lower range than before, as shown in Subfigure (b). Nevertheless, both setups are practically identical in prediction accuracy. Similar run-time improvements are observed in Subfigure (a). For nu-SVM with linear kernel, presented in Figure~\hyperref[fig:14]{14}, scaling observations are similar to the corresponding Figure~\hyperref[fig:11]{11}, which used Dataset I. Note that the prediction accuracies are only about 50\%, which is the expected value for random selection between two target classes. However, note that the main argument here is not the absolute prediction accuracy, rather closeness of it for both setups. To improve the prediction accuracy, results were obtained for a radial basis function (rbf) kernel, which is widely considered a robust kernel type in SVM classifiers, and are presented in Figure~\hyperref[fig:15]{15}. This resulted in slower learning, as observed in Subfigure (a), but near perfect prediction accuracy, as shown in Subfigure (b).

\begin{figure}[h!]
    \centering \includegraphics[width=0.5\columnwidth]{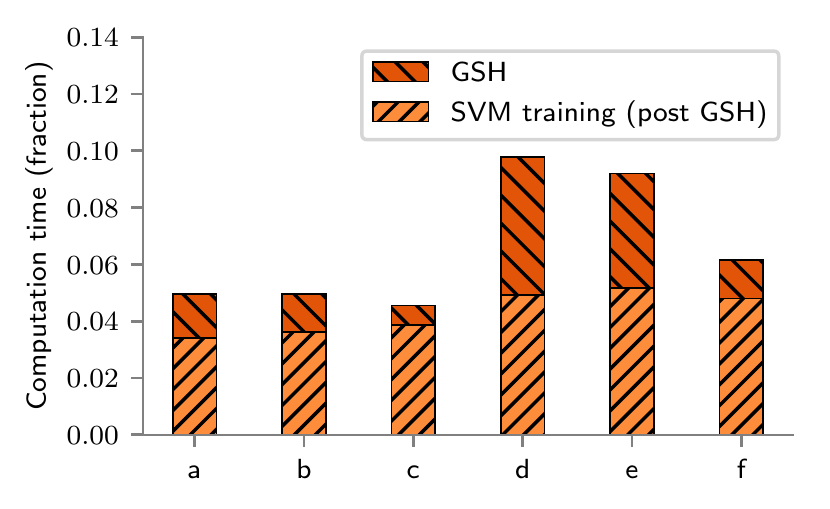}
      \caption{\csentence{Fraction of LIBSVM training time.} a, b, and c is for C-SVM, whereas d, e, and f is for nu-SVM with linear, polynomial, and rbf kernels, respectively}
      \label{fig:16}
\end{figure}

The next set of tests was conducted on Dataset III, for evaluating the performance of the heuristic on real data, and barplots in Figure~\hyperref[fig:16]{16} are used to present the findings. Computation run-time was scaled to accommodate different classification algorithms, namely C-SVM, and nu-SVM with linear, polynomial, and rbf kernels. For all six cases, reported prediction accuracy is low as shown in Table~\hyperref[tab:table_7]{7}, but close for both setups. Scaling improvements are consistent to the previous reportings for Dataset I, and II.

\begin{table}[h!]
\caption{Prediction accuracy (\%).}
\begin{tabular}{|c|c|c|c|}
  \hline
  \multicolumn{2}{|c|}{Clasification algorithm} & LIBSVM & GSH\\
  \hline
  \multirow{3}{*}{C-SVM} & linear & 58 & 58\\
  \cline{2-4}
  & polynomial & 58 & 58\\
  \cline{2-4}
  & radial & 59 & 59\\
  \hline
  \multirow{3}{*}{nu-SVM} & linear & 52 & 58\\
  \cline{2-4}
  & polynomial & 58 & 57\\
  \cline{2-4}
  & radial & 55 & 48\\
  \hline
\end{tabular}
\label{tab:table_7}
\end{table}

Overall, scalability improvements are observed consistently over a number of classifiers for Dataset I, II, and III. Furthermore, the prediction accuracy either closely follows original training or even outperforms it in a few instances.  

\subsubsection*{Serial GCH}
The next set of tests are aimed to compare run-time of serial GCH to that of GSH. More precisely, serial execution of GCH, which represents approximate learning, is fared against (reduced training data) GSH learning. In GCH, the graph clubbing step was an added cost over GSH. Results are presented for Dataset I. 

The number of iterations was set 10, and used as the termination condition for the graph clubbing algorithm. Pre-processor was used, and the collected prediction accuracies were weight-averaged over all reported partitions. Edge weight constants given in Table~\hyperref[tab:table_8]{8} were used throughout.

\begin{table}[h!]
\caption{Pattern measure constants for GCH.}
\begin{tabular}{|c|c|c|c|}
\hline
Parameter & Initial & Re-assessment\\
\hline
$C_I$ & $e$ & $e^{1.5}$\\
\hline
$C_E$ & $e^5$ & $1$\\
\hline
\end{tabular}
\label{tab:table_8}
\end{table}

\begin{figure}[h!]
    \centering \includegraphics[width=\columnwidth]{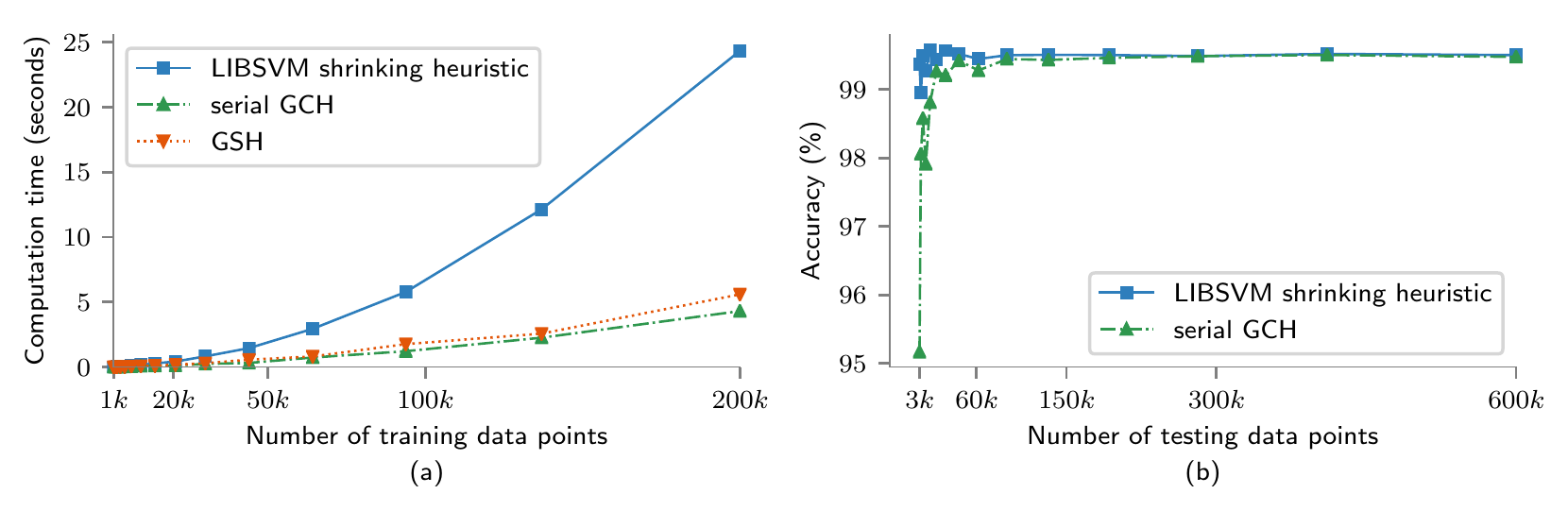}
      \caption{\csentence{C-SVM with linear kernel.}}
      \label{fig:17}
\end{figure}

\begin{figure}[h!]
    \centering \includegraphics[width=0.5\columnwidth]{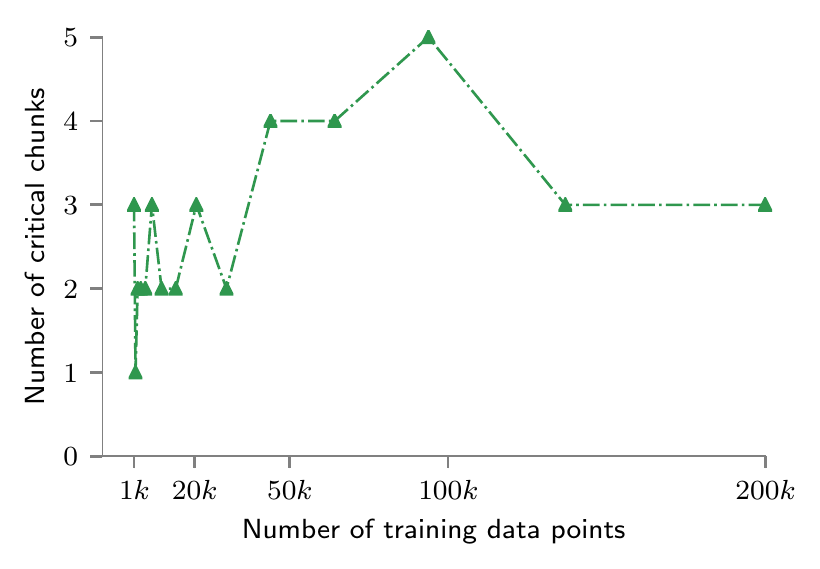}
      \caption{\csentence{Partitions after GCH.}}
      \label{fig:18}
\end{figure}

\begin{figure}[h!]
    \centering \includegraphics[width=0.5\columnwidth]{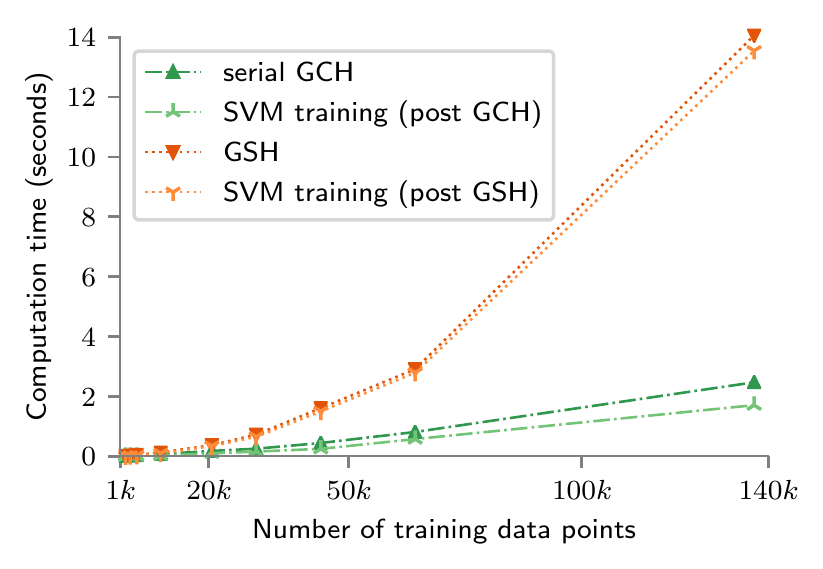}
      \caption{\csentence{C-SVM with polynomial kernel.}}
      \label{fig:19}
\end{figure}

\begin{figure}[h!]
    \centering \includegraphics[width=0.5\columnwidth]{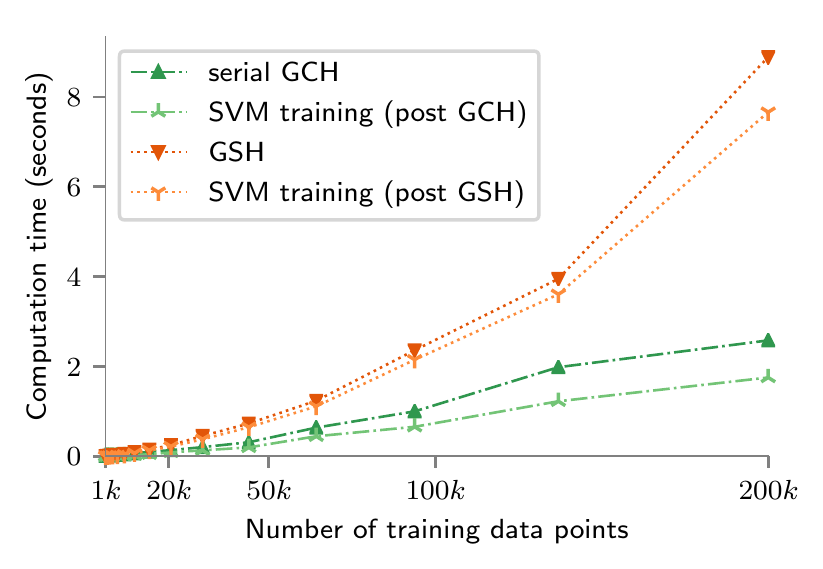}
      \caption{\csentence{nu-SVM with polynomial kernel.}}
      \label{fig:20}
\end{figure}

Only GSH, and GCH are compared along with respective line-plots of subsequent training for all the classifiers, presented in Figures~\hyperref[fig:19]{19}, and~\hyperref[fig:20]{20}. However, for the first classifier, C-SVM with linear kernel, shown in Figure~\hyperref[fig:17]{17}, computation run-time of LIBSVM with shrinking heuristic is also plotted. An accuracy plot, Subfigure (b), and a plot for the number of partitions in Figure~\hyperref[fig:18]{18} are also presented.

In Figure~\hyperref[fig:17]{17}(a), a clear scalability hierarchy is observed, with serial GCH $>$ GSH $>$ LIBSVM shrinking heuristic. In the accuracy plot, shown in Subfigure (b), approximate learning closely follows native learning for $>$8k training data points. However, for fewer points, the accuracy of the model trained after GCH, which is $\approx$95\% is not as good as native training at $99\%$. 2 to 5 numbers of partitions were obtained as shown in Figure~\hyperref[fig:18]{18}. For polynomial kernel classifiers, presented in Figure~\hyperref[fig:19]{19} -~\hyperref[fig:20]{20}, scaling advantages with serial GCH compared to GSH are observed even for lower data points ($\approx$5k).

\subsubsection*{Distributed GCH}
The main aim here was to demonstrate run-time advantage with worker processes for the communication-free training scheme discussed earlier. The scheme only aims for coarse grain parallelization, such that individual data chunks/partitions can be distributed to worker processes. 

\begin{figure}[h!]
    \centering \includegraphics[width=\columnwidth]{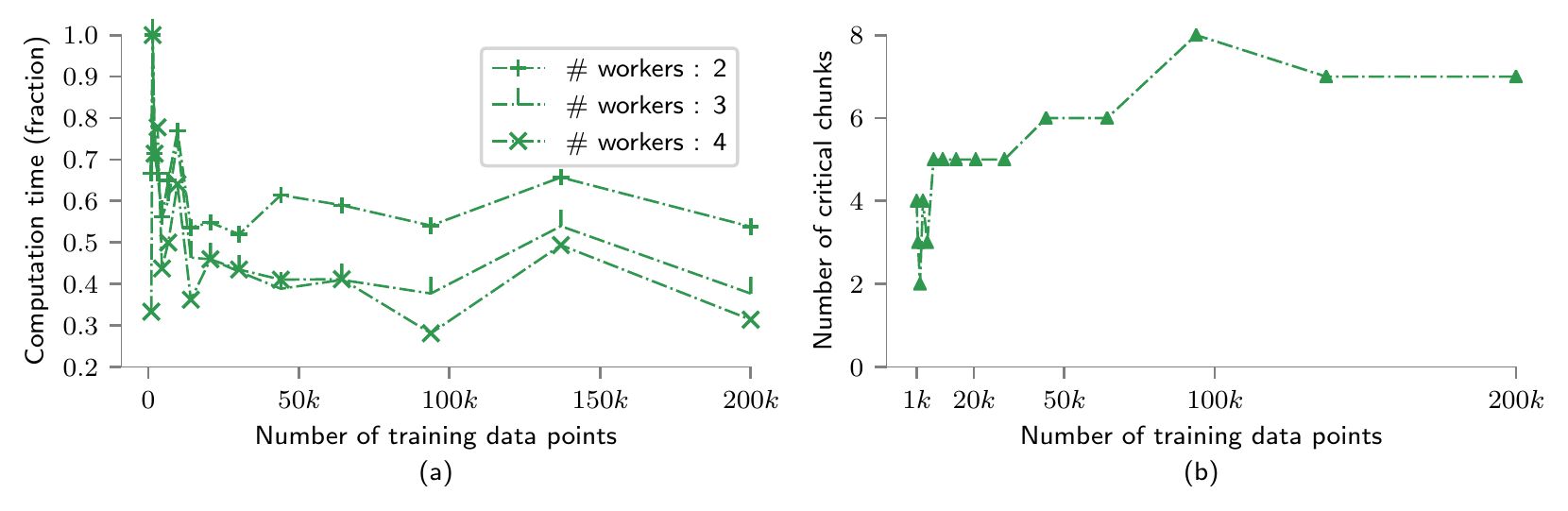}
      \caption{\csentence{Scalability with workers (a), and obtained partitions (b).}}
      \label{fig:21}
\end{figure}

Results are presented in Figure~\hyperref[fig:21]{21} for C-SVM with polynomial kernel classifier on Dataset I. The distribution invariantly depends on the number of obtained partitions. Therefore, scaling results in Subfigure (a) cannot be interpreted without the plot for the number of partitions, which is shown in Subfigure (b).

\begin{figure}[h!]
    \centering \includegraphics[width=0.5\columnwidth]{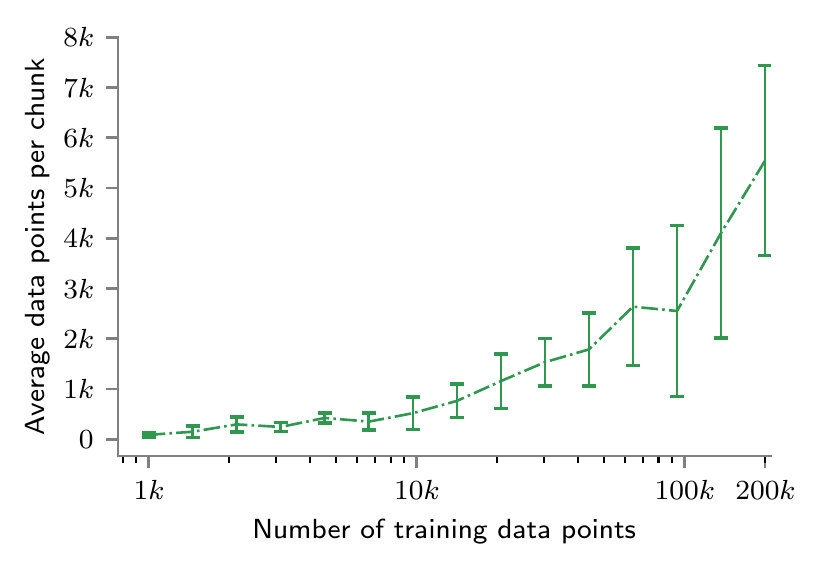}
      \caption{\csentence{Mean size of partitions with SD errorbars.}}
      \label{fig:22}
\end{figure}

Significant improvements are observed with more workers. However, sub-optimal distribution with the number of workers $>$2 can be seen in Figure~\hyperref[fig:22]{22}. Mean number of data points per chunk/partition is plotted as a function of the number of total data points. Using the number of partitions/chunks from Figure~\hyperref[fig:21]{21}(b), it is straightforward to interpret the monotonic increase of the plot. In a hypothetical case where the partitions are all equally-sized, SD would be zero. However, as the number of total data points increases, error bars for SD are observed to be commensurate to the mean number of data points. Un-equal sizes of partitions is the apparent reason for the sub-optimal distribution.

\subsubsection*{Network design}

\begin{figure}[h!]
    \centering \includegraphics[width=0.5\columnwidth]{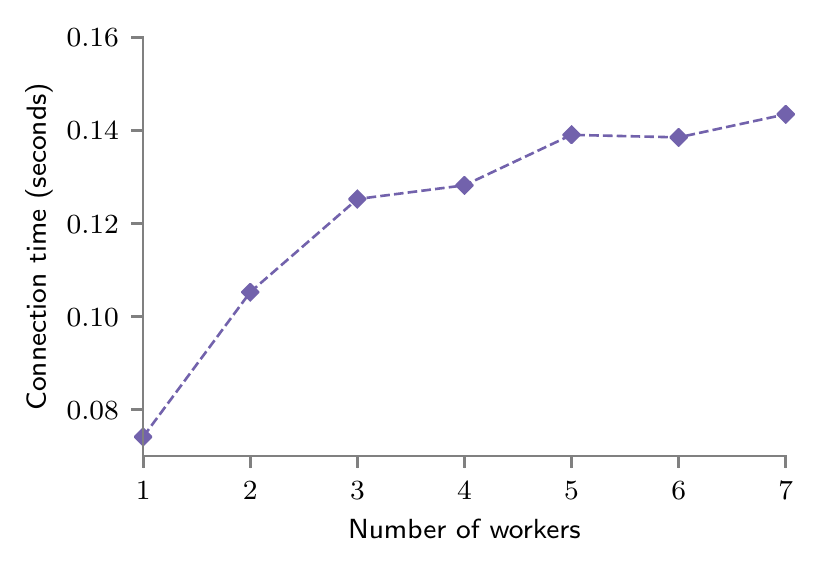}
      \caption{\csentence{Performance of the designed connection procedure}}
      \label{fig:23}
\end{figure}

The main motivation for this set of tests was to evaluate various implementation aspects of the network that implemented distributed GCH. The first aspect is the connection time of the distributed application. Figure~\hyperref[fig:23]{23} shows connection time measurements with the number of workers. 

Apparently, the connection time is very small ($\approx$120 ms) when compared to the run-time of the training phase. The connection procedure was replicated on another distributed messaging API, MPICH3.3 \cite{mpi}, and the connection time with ZeroMQ was observed to be 1.53 times faster than with MPICH3.3.

\begin{figure}[h!]
    \centering \includegraphics[width=0.5\columnwidth]{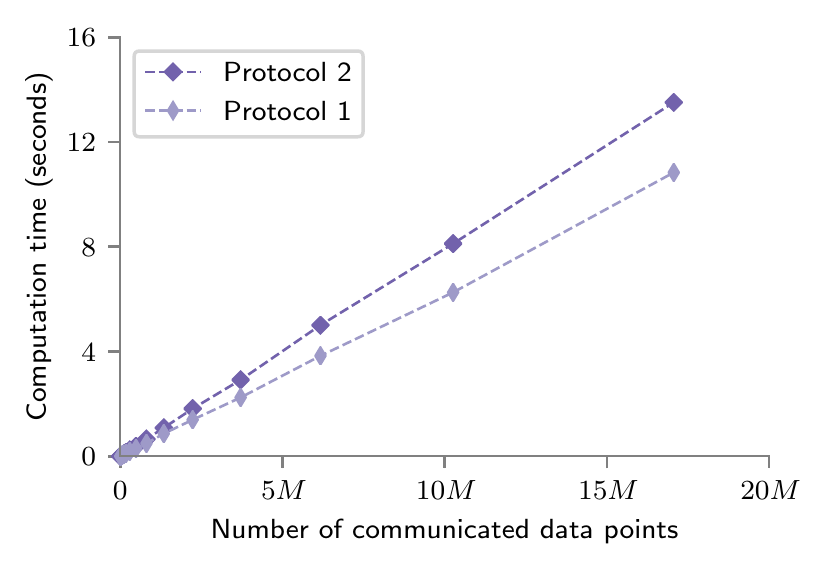}
      \caption{\csentence{Evaluation of messaging protocols.} Protocol 1 is used in this work.}
      \label{fig:24}
\end{figure}

The next aspect of the network is the messaging protocols. The primary motivation behind the protocol design is to utilize the shape of the \ensuremath{i^{\text{th}}} data point, that has n-dimensional feature vector ($x_{i}$), and 1-dimensional target class value (tc\ensuremath{(i))}. Two of such protocols as previously discussed were fared in run-time. A couple of opposing factors are at play between these protocols. Data marshaling is an extra step in protocol 1 (before actual messaging) when compared to protocol 2. On the other hand, message count/traffic increases (multiple times governed by the dimensionality of data) in protocol 2 compared to protocol 1. Overall, protocol 1 performed better as can be been in Figure~\hyperref[fig:24]{24}. Although increased marshaling may probably result in even better scaling, a larger memory footprint on the channel is un-advisable as it can result in a overflow scenario.

\begin{figure}[h!]
    \centering \includegraphics[width=0.5\columnwidth]{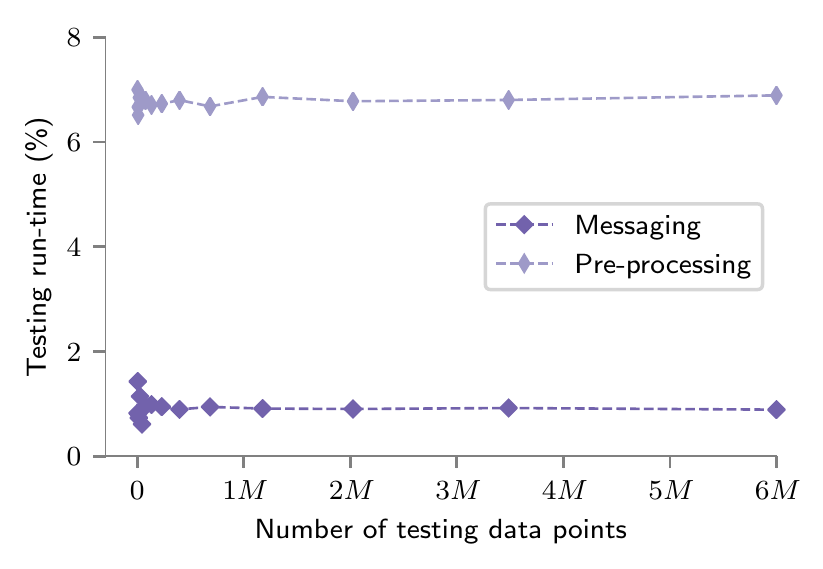}
      \caption{\csentence{Messaging, and pre-processing.} Run-time of message communications, and pre-processing measured in percentage of the testing phase.}
      \label{fig:25}
\end{figure}

Messaging efficiency was tested in the context of prediction/testing phase of the distributed implementation. Reportedly in Figure~\hyperref[fig:25]{25}, it is very efficient at $\approx$1\% of the prediction time. It is found to be even more efficient at $\approx$0.5\% for the training phase. Pre-processing run-time is at $\approx$7\%, as can be seen in Figure~\hyperref[fig:25]{25}.

Note that external libraries are used in the formulution, including a clustering method, an approximate nearest neighbor implementation, two messaging APIs, and many classification algorithms. The heuristic execution is not very sensitive to the parameters of these external implementations. For instance, only 3~-~5 iterations of clustering were needed, input parameters for approximate nearest neighbor, and classification algorithms were kept at default. However, the heuristic is sensitive to the parameters which are part of the heuristic steps.

\section*{Conclusions}
In this work, a three parts algorithm, and an edge weight scheme are designed to construct a weighted graph that effectively captures the classification patterns of dataset. Another three part graph coarsening algorithm with the directional aspect is designed that divides the reduced dataset into partitions that can be trained independently using the designed communication-free network.

GSH provides an evident serial scalability advantage, and generic applicability holds for every classifier. It encourages to extrapolate the argument, and claim that GSH will hold for a multitude of classification algorithms. Approximate learning is the reason behind serial GCH's added advantage over GSH. A better model training throughput can be achieved via the use of both of these approaches.

Distribution of chunks/partitions of the training set to worker processes results in further scaling improvements. Even for a nominal number of worker processes (say four), available in almost all current computing platforms, parallel performance benefits can be achieved. However, un-equal coarsening, and control over the number of partitions still needs to be further investigated in the current implementation of GCH. 

First, a possible direction of investigation is the directional aspect of the objective. That is, if the partitioning was better directed, better orthogonality between the contour of the underlying classification pattern, and of partition boundaries shall be obtained. This condition is necessary for correctness (in prediction accuracy) of this approximate procedure. The two-fold pattern measure scheme should be designed to get the desired control over the direction of coarsening.

A second possible investigation is modifying the objective function of the current graph coarsening scheme. Part of traditional coarsening objectives can be added to the current objective. Since popular objectives predominantly optimized size division of partitions, it will address the un-equal coarsening problem in the presented objective. Furthermore, the number of partitions is directly controlled in these objectives.

All approaches of the presented heuristic, namely GSH, serial, and distributed GCH, are effective to reduce training run-time of a classification algorithm. Furthermore, scaling benefits are accompanied with no compromise in prediction accuracy.

\printacronyms[name=List of Abbreviations, include-classes={abbrev, nomencl}, sort = false]

\begin{backmatter}

\section*{Declarations}

\section*{Availability of data and materials}

Dataset I, and II used during the current study are available from the corresponding author on a reasonable request. Furthermore, two real datasets used are available in the UCI machine learning repository, https://archive.ics.uci.edu/ml/datasets/skin+segmentation; https://archive.ics.uci.edu/ml/datasets/Cuff-Less+Blood+Pressure+Estimation. Other resources with respect to the manuscript are shared at \url{https://sumedhyadav.github.io/projects/graph_heuristic}.

Tests were performed with following computing specfications:
\begin{itemize}
  \item Operating system(s): Ubuntu 16.04 LTS
    \item Programming language: C/C++, python (for visualization)
    \item Visualization tools: matplotlib, and networkx modules in python
    \item External Libraries: FLANN, mlpack, LIBSVM
    \item Version requirements: g++ 5.4.0, Python 2.7.12, matplotlib 1.5.1, networkx 2.2
\end{itemize}

\section*{Competing interests}
The authors declare that they have no competing interests.

\section*{Funding}
Not applicable

\section*{Author's contributions}
This work is joint research of SY and MB. SY did all implementation, and run all test cases. All authors read, and approved the final manuscript.

\section*{Acknowledgements}
SY acknowledges Gautam Kumar's help with respect to preparing the manuscript.

\section*{Authors' information}
Not applicable

\end{backmatter}
\end{document}